%% file: arxiv.tex
\documentclass[10pt,twocolumn,letterpaper]{article}

\usepackage[pagenumbers]{preprint}

\input{preamble}

\usepackage{tabularx}
\usepackage[normalem]{ulem}
\usepackage{adjustbox}
\newsavebox{\fittab}
\usepackage{placeins}            %
\usepackage{newunicodechar}
\newunicodechar{±}{\ensuremath{\pm}}
\newunicodechar{−}{\ensuremath{-}}
\newunicodechar{–}{--}
\newunicodechar{—}{---}
\newunicodechar{▸}{\ensuremath{\blacktriangleright}}
\newunicodechar{◆}{\ensuremath{\blacklozenge}}
\newunicodechar{↔}{\ensuremath{\leftrightarrow}}
\newunicodechar{→}{\ensuremath{\rightarrow}}
\newunicodechar{↑}{\ensuremath{\uparrow}}
\newunicodechar{↓}{\ensuremath{\downarrow}}
\newunicodechar{†}{\dag}
\newunicodechar{‡}{\ddag}
\newunicodechar{⚑}{\shortcutflag}
\newunicodechar{‖}{\ensuremath{\|}}
\newunicodechar{×}{\ensuremath{\times}}
\newunicodechar{²}{\textsuperscript{2}}
\newunicodechar{≥}{\ensuremath{\geq}}
\newunicodechar{≤}{\ensuremath{\leq}}
\newunicodechar{≈}{\ensuremath{\approx}}
\newunicodechar{ρ}{\ensuremath{\rho}}
\newunicodechar{λ}{\ensuremath{\lambda}}
\newunicodechar{τ}{\ensuremath{\tau}}
\newunicodechar{Σ}{\ensuremath{\Sigma}}
\newunicodechar{ₜ}{\ensuremath{_{t}}}
\newunicodechar{ⱼ}{\ensuremath{_{j}}}

\definecolor{linkblue}{rgb}{0.21,0.49,0.74}
\usepackage[pagebackref,breaklinks,colorlinks,allcolors=linkblue]{hyperref}

\title{\method: Explaining Frozen Foundation Encoder-based Medical Classification by Exact Differential Diagnosis}

\author{Luojie Huang, Gregory N. McKay, Nicholas J. Durr$^{*}$\\
Department of Biomedical Engineering, Johns Hopkins University, Baltimore, USA\\
{\tt\small \{lhuang48, gmckay1, ndurr\}@jhu.edu}}

\begin{document}

\twocolumn[{%
\renewcommand\twocolumn[1][]{#1}%
\maketitle
\begin{center}
  \centering
  \captionsetup{type=figure}
  \includegraphics[width=\textwidth]{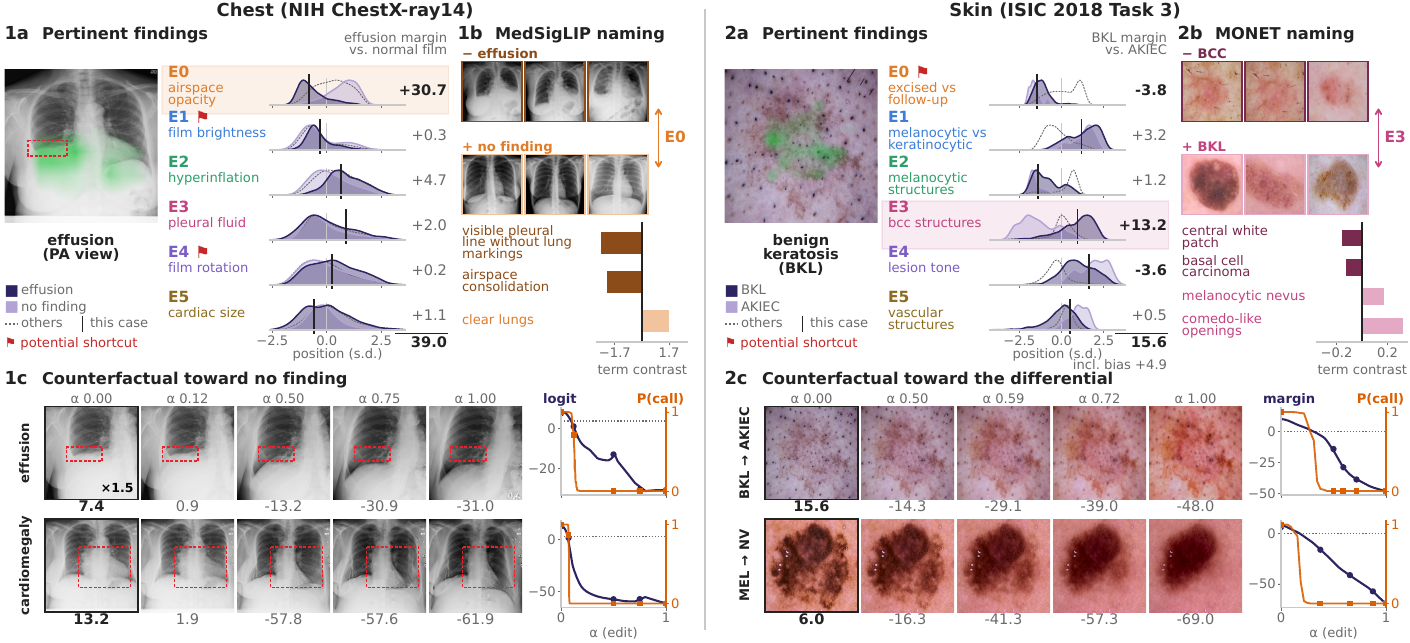}
  \captionof{figure}{\textbf{Differential diagnosis from \method.}
  A chest film called effusion (left) and a skin lesion called benign keratosis (BKL) against actinic keratosis (AKIEC) (right).
  (1a, 2a) Pertinent findings. Each axis places the case among the class distributions, and its evidence against the alternative sums exactly to the decision margin (with the class bias on skin). Green maps the deciding axis (shaded row), dashed red is the NIH box, and \shortcutflag{} marks a potential shortcut axis.
  (1b, 2b) Pole naming with a medical VLM.
  (1c, 2c) Counterfactual edits. The framed case moves a fraction $\alpha$ toward the alternative prototype and is classified again until the prediction flips (dotted line).}
  \label{fig:teaser}
\end{center}
\vspace{0.5em}
}]

{\let\thefootnote\relax\begin{NoHyper}\footnotetext{$^{*}$Corresponding author: Nicholas J. Durr (ndurr@jhu.edu).}\end{NoHyper}}
\input{sec/0_abstract}
\input{sec/1_intro}
\input{sec/2_related}
\input{sec/3_method}
\input{sec/4_experiments}

\input{sec/5_results}
\input{sec/6_conclusion}

\section*{Acknowledgment}
This work was supported by the Gates Foundation [\mbox{INV-077759} and \mbox{INV-095603}]. The conclusions and opinions expressed in this work are those of the authors alone and shall not be attributed to the Foundation.

{
    \small
    \phantomsection\label{sec:refs}
    \bibliographystyle{ieeenat_fullname_unsrt}
    \bibliography{references}
}

\setlength{\textfloatsep}{12pt plus 2pt minus 2pt}\setlength{\dbltextfloatsep}{12pt plus 2pt minus 2pt}
\raggedbottom
\makeatletter\setlength{\@fptop}{0pt}\setlength{\@dblfptop}{0pt}\makeatother
\setcounter{dbltopnumber}{3}\renewcommand{\dbltopfraction}{0.95}\renewcommand{\dblfloatpagefraction}{0.5}
\let\ddxOrigSection\section
\renewcommand{\section}{\FloatBarrier\ddxOrigSection}
\renewcommand{\theHsection}{S\arabic{section}}
\input{sec/X_suppl_arxiv}

\end{document}

%% file: preamble.tex
\usepackage{microtype}

\renewcommand{\paragraph}[1]{\vspace{.5em}\noindent\textbf{#1}} %

\usepackage{xspace}
\newcommand{\method}{DDxBasis\xspace}

\usepackage{iftex}
\usepackage{colortbl}                  %
\usepackage{makecell}                  %
\definecolor{rowshade}{HTML}{E7F1FB}   %
\usepackage{fontawesome5}              %
\newcommand{\frozenicon}{\faSnowflake[regular]}
\newcommand{\trainedicon}{\faFire}
\newcommand{\newicon}{\faStar}
\newcommand{\shortcutflag}{\faFlag}
\newcommand{\piup}[1][\pm]{\pi^{#1}_{uj}}

%% file: sec/0_abstract.tex
\begin{abstract}
In differential diagnosis (DDx), a clinician explains a diagnosis by the few findings that separate it from the alternatives.
Medical image classifiers cannot yet explain themselves this way.
Existing explanations are either faithful but spread over hundreds of concepts, or compact but leave part of each decision unexplained.
We present \method, which explains a classifier on a frozen foundation encoder as a differential diagnosis in three steps.
First, a concept bottleneck reads all patch tokens at every encoder layer using gated, second-order slot attention.
A concept basis then regroups the concepts into two-pole axes, each a differential between groups of diagnoses.
Finally, we read a differential diagnosis from each axis through the typical presentation of its poles, the pertinent findings of the case, and counterfactual edits of borderline cases.
Across five medical datasets, \method is the most accurate frozen-encoder model and close to the best published domain-specific results.
Because every concept adds its own evidence, its reproducible concept basis splits each decision exactly, at no cost in accuracy.
Without ordered stage or concept labels, the axes recover cell lineages and rare subtypes, agree with dermatologists' structure annotations, and expose planted and real shortcuts.
Project page: \mbox{\href{https://durrlab.github.io/DDxBasis/}{\texttt{durrlab.github.io/DDxBasis}}}
\end{abstract}

%% file: sec/1_intro.tex
\section{Introduction}
\label{sec:intro}

Foundation encoders give promising zero-shot~\cite{BiomedCLIP,MedGemma} and few-shot~\cite{Shakeri2024} medical classification, from vision--language models (VLMs) to self-supervised encoders such as DINOv2 and DINOv3~\cite{DINOv2,DINOv3}.
However, accuracy alone is not enough for clinicians to act on a prediction.
Clinicians reason by differential diagnosis (DDx), weighing the findings that separate each candidate diagnosis from its competitors~\cite{Bowen2006}.
Previous AI systems return such differentials as ranked lists~\cite{AMIE}, but not the reasoning behind the rank.
For a benign keratosis that the model nearly calls an actinic keratosis (\cref{fig:teaser}, right), the dermatologist needs to know which findings support each diagnosis.
Such contrastive explanations are clinically useful~\cite{Miller2019} and trusted~\cite{Chanda2024}, but they must be faithful, since a plausible explanation of a wrong prediction misleads even experts~\cite{Tschandl2020}.

Saliency maps~\cite{GradCAM} show where evidence lies but not what it is~\cite{Adebayo2018}, and post-hoc counterfactuals~\cite{DiME} visualize what change in the image might flip the decision but not how the model reads it.
Prototype networks~\cite{ProtoPNet,PIPNet} and concept bottlenecks~\cite{CBM,BotCL} are faithful by construction, but over a hundred overlapping concepts are too many to weigh.
Post-hoc dictionaries~\cite{CRAFT,Bricken2023} are compact but leave an unexplained residual~\cite{IBD,ConceptSHAP}.

We present \method, which obtains faithful and compact explanations by regrouping an accurate concept bottleneck.
Our concept bottleneck achieves strong classification by reading all patch tokens across every layer of the frozen encoder with competing, gated, second-order slots (\cref{sec:bottleneck}).
Each class logit is a sum of non-negative concept evidence.
Summing the same concept evidence group by group gives the same class logits, so an explanation by concept groups is exact and faithful to the model.
To read the grouped concepts for differential diagnosis, we propose a compact set of novel two-pole axes, named a \emph{concept basis}, each a differential between groups of diagnoses, trained to be distinct, balanced, contrastive, and reproducible (\cref{sec:basis}).

Our contributions are threefold.
First, our concept bottleneck turns all patch tokens of a frozen foundation encoder into a strong classifier.
On five medical tasks in dermoscopy, blood and bone-marrow cytology, phase-contrast cell microscopy, and chest radiography, \method is the most accurate frozen-encoder model, on average 12 points above a linear probe on the same encoder, and close to published domain-specific results (\cref{sec:acc}).
Second, the explanation is both exact and compact.
Each decision rests on a few reproducible axes of the concept basis, from which we read a differential diagnosis through the typical presentation of their poles (\cref{fig:teaser}, 1b and 2b), the pertinent findings of the case (\cref{fig:teaser}, 1a and 2a), and counterfactual edits of borderline cases (\cref{fig:teaser}, 1c and 2c).
Third, the axes reveal insights beyond class labels, from clinical structure to shortcuts.
Without ordered stage or concept labels, the axes discover cell lineages (one axis orders five granulocyte maturation stages with Spearman $\rho$ 0.87 from only two merged training labels), flag rare subtypes hidden inside coarse training classes, and follow the dermoscopic structures verified by dermatologists' annotations (\cref{sec:clinical}).
They also expose shortcuts, cues that predict the label in training but not the disease~\cite{Geirhos2020}.
A planted shortcut lands on one axis whose removal raises accuracy by about 20 points where the strong shortcut misleads, and the view position of chest radiographs is flagged as a real shortcut (\cref{sec:audit}).

%% file: sec/2_related.tex
\section{Related Work}
\label{sec:related}

\begin{figure*}[t]
  \centering
  \includegraphics[width=\textwidth]{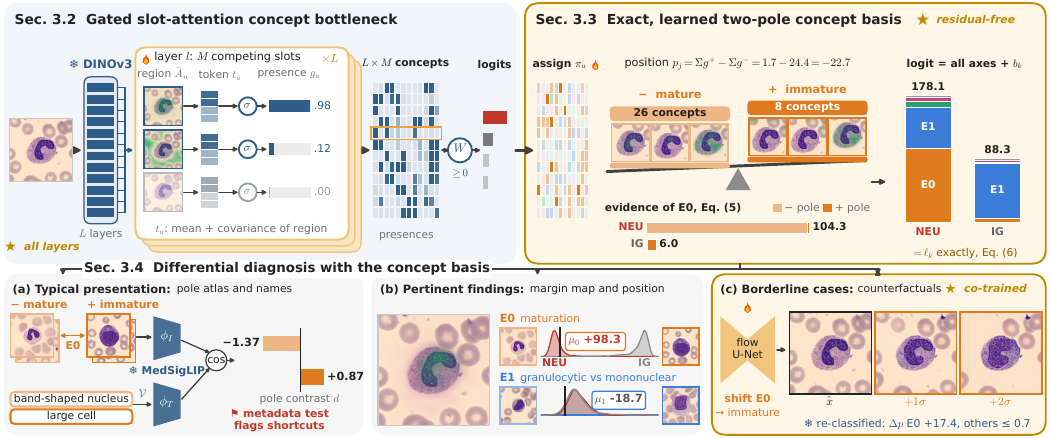}
  \caption{\textbf{Overview of \method} on a PBC band neutrophil.
  Gated slots on every layer detect concepts (\cref{sec:bottleneck}), a two-pole basis regroups them into axes whose evidence sums exactly to the logits (\cref{sec:basis}), and each axis is read via (a) pole atlases with VLM names, (b) its margin map, position, and margin, and (c) counterfactual edits toward the alternative (\cref{sec:reading}).
  \frozenicon{} frozen; \trainedicon{} trained; \newicon{} new.}
  \label{fig:pipeline}
\end{figure*}

\paragraph{Foundation models for medical images.}
Frozen foundation encoders, from self-supervised vision models~\cite{DINOv2,DINOv3} to VLMs~\cite{CLIP,BiomedCLIP,MedGemma}, transfer well to medical classification~\cite{Huix2024,Baharoon2023,Liu2025}, segmentation~\cite{Liu2025}, and report generation~\cite{MedGemma}.
DINOv3, trained only on natural images, even outperforms medical-specific encoders on several tasks~\cite{Liu2025}, although encoders pretrained on pathology~\cite{UNI}, retina~\cite{RETFound}, blood cells~\cite{DinoBloom}, or chest radiographs~\cite{RayDINO} still lead where the domain gap is large.
With few labels per task, adaptations range from last-layer probes~\cite{DINOv2,AIM} and cross-layer fusion~\cite{ELMo,Head2Toe,ALF} to parameter-efficient fine-tuning~\cite{LoRA}.
Previous work reported that intermediate layers transfer as well as the last~\cite{Head2Toe}, but cross-layer readouts fuse them into one feature, whereas we sum the evidence of per-layer regional concepts.

\paragraph{Concept bottlenecks and prototypes.}
A concept bottleneck model (CBM) predicts annotated concepts and then the class~\cite{CBM}, but such annotations are scarce in medicine, and soft concept scores can leak class information~\cite{Havasi2022}.
Label-free and post-hoc CBMs score concepts with VLMs~\cite{LFCBM,PCBM}, name a sparse dictionary~\cite{DNCBM}, or match diagnostic criteria written in text~\cite{Explicd}, so their concepts are limited to vocabulary banks.
Prototype networks learn concepts from the images instead~\cite{ProtoPNet,PIPNet}.
Their variants share prototypes~\cite{ProtoPShare}, use transformers~\cite{ProtoViT} or foundation features~\cite{PNP}, span several feature levels~\cite{MCPNet}, or explain chest radiographs~\cite{XProtoNet}, but most read one layer through over a hundred prototypes.
In BotCL~\cite{BotCL} and MUFASA~\cite{MUFASA}, slots compete for every patch~\cite{SlotAttention}, so context and acquisition cues also receive concepts, but BotCL reads one layer and MUFASA fuses several layers into one slot set.
We extend competing slots to every layer and keep each layer's evidence separate, so that the concepts can be regrouped exactly.

\paragraph{Post-hoc concepts and contrastive axes.}
TCAV~\cite{TCAV}, CRAFT~\cite{CRAFT}, and sparse autoencoders~\cite{Bricken2023} explain a fixed model with concept directions or dictionaries.
Two-pole scales appear as hand-picked axes in the semantic differential~\cite{Osgood1957}, SemAxis~\cite{SemAxis}, and relative TCAV~\cite{TCAV}, exact decompositions come from additive concept heads~\cite{PIPNet}, and concept groups from clustering or factorization~\cite{Vigneau2003,CRAFT}.
Rotations and whitening~\cite{Kaiser1958,ConceptWhitening} decorrelate only by giving up partition or exactness, and independence fails for correlated factors~\cite{Trauble2021}.
In medical imaging, counterfactual generators show what changes a prediction and can expose shortcuts~\cite{LatentShift,Singla2023,FastDiME}, and concept bottleneck generative models steer generation through supervised concepts~\cite{CBGM}.
Our concept basis learns two-pole axes with a log-determinant volume for distinctness, and conditions a co-trained generator for counterfactual edits of one axis at a time.

\paragraph{Medical interpretability.}
Medical concept models build clinical criteria in before training, from clinical notes or literature~\cite{Explicd,MONET}, while prototype models compare a case with learned disease prototypes~\cite{XProtoNet}.
Medical classifiers also learn shortcuts, such as radiograph projection and acquisition artifacts~\cite{Zech2018,Bissoto2019,Winkler2019,Pereira2024}, and fail on rare subtypes~\cite{OakdenRayner2020}.
Slice-discovery methods search a model's errors for such failure groups~\cite{Domino}.
Our axes are learned without concept labels and named post hoc with validated terms, so they are not bound to predefined criteria and can reveal shortcuts, hidden subtypes, and the clinical differential behind errors.

%% file: sec/3_method.tex
\section{Method}
\label{sec:method}

\paragraph{Overview (\cref{fig:pipeline}).}
\method has three parts: a concept bottleneck reads every layer of a frozen foundation encoder with gated slot attention (\cref{sec:bottleneck}), a novel concept basis regroups its concepts exactly into $K$ two-pole axes (\cref{sec:basis}), and each axis is read through margin maps, pole atlases, VLM-named terms, and counterfactual edits (\cref{sec:reading}).
No concept labels are used, and no imaging modality is assumed.

\paragraph{Notation.}
An image $x$ has a label $y$ over $C$ classes indexed by $k$, or $y\in\{0,1\}^{C}$ for multi-label tasks.
Concepts are indexed by $u$ and axes by $j\in\{1,\dots,K\}$, and $\Delta^{n}$ is the probability simplex.

\subsection{Preliminaries: a frozen foundation encoder}
\label{sec:prelim}

We read a frozen self-supervised Vision Transformer~\cite{ViT} $f$ with $L$ layers, such as DINOv3~\cite{DINOv3}.
As intermediate layers transfer as well as the last to medical tasks~\cite{Head2Toe}, we take the patch tokens of every layer after the final LayerNorm,
\begin{equation}
  H^{(l)}(x)=\mathrm{LN}_f\big(Z^{(l)}(x)\big)_{\text{patch}}\in\mathbb{R}^{N\times D},\qquad l=1,\dots,L,
  \label{eq:encoder}
\end{equation}
where $Z^{(l)}(x)$ is the output of layer $l$ and $\mathrm{LN}_f$ the final LayerNorm of $f$, so each layer gives $N$ patch tokens $H_n$.

\subsection{A gated slot-attention concept bottleneck}
\label{sec:bottleneck}

Class evidence spreads over the whole image, from the lesion and its border to the background and acquisition artifacts.
We therefore build our concept bottleneck on slot attention~\cite{SlotAttention}, whose slots compete for every patch, and adapt it to classification in three ways: slots at every depth, second-order region descriptors, and gated non-negative evidence summed exactly over depth.
A concept here is a learned unit found without annotations, as in BotCL~\cite{BotCL}.

\paragraph{Competing slots at every depth.}
At each layer, $M$ slots compete for the patch tokens through slot attention and section the image into a few regions~\cite{DINOSAUR}.
The attention of slot $u$ renormalized over patches, $\bar A_u(x)\in\Delta^{N}$, is the region of concept $u$.
In addition to the region mean used by previous slot-based concept models~\cite{BotCL}, we introduce covariance~\cite{MPNCOV} to each of our $U=LM$ concepts, which keeps diagnostic texture, such as lesion border symmetry or a chromatin pattern, that a class token or mean averages away,
\begin{equation}
  \begin{gathered}
    t_u=s_u+F_l\,\mathrm{vech}\big(\Sigma_u^{1/2}\big),\\
    \Sigma_u=\textstyle\sum_n \bar A_{u,n}\,(V' H_n-\mu'_u)(V' H_n-\mu'_u)^{\!\top},
  \end{gathered}
  \label{eq:token}
\end{equation}
where $t_u$ is the token of concept $u$, $s_u$ its final slot state, $V'$ a linear projection, $\mu'_u$ the region mean, $\mathrm{vech}$ the upper triangle, and $F_l$ a per-layer norm and linear layer.

\paragraph{Presence gate and non-negative head.}
Each concept learns a \emph{presence} $g_u(x)=\sigma\big(P_u t_u(x)+\beta_u\big)\in(0,1)$, with a linear readout $P_u$ and bias $\beta_u$, that states whether its region shows its pattern.
The presence adds the concept evidence $e_{uk}(x)=g_u(x)W_{uk}$ to class $k$ through a weight $W_{uk}\geq 0$~\cite{PIPNet}, so layer $l$ scores $\ell^{(l)}_k(x)=\sum_u g_u(x)W_{uk}+b^{(l)}_k$ over its $M$ concepts, with class bias $b^{(l)}_k$.

\paragraph{Exact sum across layers.}
Each layer has its own slots, head, and loss, as in deep supervision~\cite{DSN}.
The layers are merged by a plain sum to prevent overfitting and make each concept's share of a logit explicit,
\begin{equation}
  \ell_k(x)=\sum_{l=1}^{L}\ell^{(l)}_k(x)=\sum_{u=1}^{U} g_u(x)W_{uk}+\sum_{l=1}^{L}b^{(l)}_k .
  \label{eq:sum}
\end{equation}

\paragraph{Concept bottleneck training.}
Each layer reconstructs patch tokens from its slots with a broadcast decoder~\cite{DINOSAUR} and classifies with a logit-adjusted cross-entropy~\cite{Menon2021} and an $\ell_1$ penalty on $W$.
A class-entropy hinge keeps each concept from serving a single class, so it can enter differentials.

\subsection{An exact, learned two-pole concept basis}
\label{sec:basis}

The bottleneck is faithful, but a decision spreads over more than a hundred overlapping concepts.
We regroup them exactly into $K\ll U$ two-pole axes, each a graded scale that contrasts two groups of classes.

\paragraph{Two-pole assignment.}
Each concept is assigned to the $2K$ poles by a learned sparse distribution
\begin{equation}
  \pi_u=1.5\text{-entmax}\big(\theta_u/\tau\big)\in\Delta^{2K},
  \label{eq:entmax}
\end{equation}
with parameters $\theta_u$ and a temperature $\tau$ annealed from 1 to $\tau_{\min}$.
The 1.5-entmax gives exact zeros~\cite{Peters2019}, so that most concepts commit to a single pole.
We write $\piup$ for the mass of concept $u$ on the two poles of axis $j$.

\paragraph{Exact decomposition.}
The evidence of axis $j$ for class $k$ pools the concepts on its poles,
\begin{equation}
  s_{jk}(x)=\sum_u \big(\pi^{+}_{uj}+\pi^{-}_{uj}\big)\,g_u(x)W_{uk},
  \label{eq:axisev}
\end{equation}
and since every $\pi_u$ sums to one, the axes split the logits of \cref{eq:sum} exactly,
\begin{equation}
  \sum_{j=1}^{K} s_{jk}(x)+\sum_{l=1}^{L}b^{(l)}_k=\ell_k(x)\qquad\forall\,x,k,
  \label{eq:exact}
\end{equation}
with no residual, unlike post-hoc decompositions~\cite{IBD,ConceptSHAP}, so the basis keeps the accuracy of the bottleneck.

\paragraph{Positions and margins.}
Each axis is a bipolar scale~\cite{SemAxis}, on which the \emph{position} of $x$ is its presence on the positive pole minus that on the negative pole,
\begin{equation}
  p_j(x)=\sum_u\big(\pi^{+}_{uj}-\pi^{-}_{uj}\big)\,g_u(x),
  \label{eq:position}
\end{equation}
independent of any class.
Exactness holds for the evidence $s_{jk}$ over all $U$ concepts, whereas $p_j$ places an image on the axis.
The construction pairs two groups of concepts on one signed scale.
The pole-coherence term below encourages concepts on opposite poles to vary in opposite directions within a class, so that the position measures where an image lies between them.
For a prediction $\hat k$ and the class it is weighed against, $\tilde k$, each axis contributes a margin $\mu_j=s_{j\hat k}-s_{j\tilde k}$, and these margins also sum exactly to the logit margin.
Evidence belongs to one class, whereas a margin compares evidence between two classes.

\paragraph{Learning the basis.}
Four regularization terms prevent the concept basis from collapsing onto a few dominant traits (exact forms in Supp.\ Sec.~S1).
A log-determinant volume~\cite{Kulesza2012} on the within-class correlation of positions keeps the axes \emph{distinct}, penalizing collinear axes sharply while staying flat near orthogonality, allowing the mild correlation of co-occurring clinical attributes~\cite{Trauble2021}.
A load-balancing band keeps them \emph{balanced}, holding each axis's average share of the margin between $\kappa_{\min}$ and $\kappa_{\max}$.
A normalized pole score on the within-class correlation of presences makes the poles \emph{coherent}, so that presences correlate positively within a pole and negatively across poles.
Finally, bottlenecks trained on the same dataset with different seeds stably section an image into similar regions (\cref{tab:ablation}a), so the concept basis inherits this \emph{reproducibility} when we learn the bases of $S$ bottlenecks jointly from one PCA--varimax start on the within-class presences~\cite{Kaiser1958}, and a consensus term ties matching axes across bottlenecks to the same image property.

\subsection{Differential diagnosis with the concept basis}
\label{sec:reading}

A differential diagnosis is read from the findings that discriminate its candidates.
The clinician recalls the typical presentation of each candidate, locates and weighs the pertinent positive and negative findings in the case, and, in a borderline case, considers how it would present closer to the alternative.
Each two-pole axis states one such differential contrast, and we read the axes with the largest margins $\mu_j$ (\cref{sec:basis}) in these three steps.

\paragraph{Typical presentation: pole atlas and names, \cref{fig:teaser} (1b, 2b).}
The pole atlas presents each pole by the confidently and correctly classified images with the highest pole score $\sum_u\big(\piup-\tfrac12\piup[\mp]\big)g_u(x)$, along with pole maps $\mathcal{P}^{\pm}_j=\sum_u\piup g_u\bar A_u$.
To name a pole, we use pretrained medical VLMs, MONET for dermatology~\cite{MONET} and MedSigLIP otherwise~\cite{MedGemma}, to score a set of clinical terms against training images~\cite{DNCBM}.
Since the two poles favor different diagnoses, we measure the term contrast by comparing, within each class, the mean scores of the images furthest toward each pole, so that a term names a finding rather than a class.
A term is kept only if the difference is significant and replicates on validation images.
Each axis is labeled from dataset labels and metadata where they suffice, and otherwise from its pole names (Supp.\ Sec.~S2).

\paragraph{Pertinent findings: margin map and position, \cref{fig:teaser} (1a, 2a).}
The margin map $\mathcal{M}_j(x)=\sum_u(\piup[+]+\piup[-])\big(e_{u\hat k}(x)-e_{u\tilde k}(x)\big)\bar A_u(x)$, with the concept evidence $e_{uk}$, spreads each concept's contrast over its attention and sums exactly to the axis margin $\mu_j$.
Its positive regions are the pertinent findings for the prediction and its negative regions those for the alternative~\cite{CEM}, and the position $p_j(x)$, read against the candidates' within-class distributions, shows which candidate diagnosis the case leans toward.

\paragraph{Borderline cases: counterfactual edits, \cref{fig:teaser} (1c, 2c).}
We render how \emph{this} case would present further toward the alternative~\cite{DiME}.
A pixel-space rectified flow~\cite{RectifiedFlow}, co-trained with the basis, is conditioned by cross-attention~\cite{LDM} on one token per axis, built from its position $p_j$ and content code, a signed average of its concept tokens $t_u$ (Supp.\ Sec.~S3), and a descriptor residual token for general context with capped label-explained variance.
The generator reconstructs the image faithfully, and an axis edit changes the finding rather than unrelated properties (Supp.\ Table~S11).
To edit, we invert the image to noise under its condition, shift one axis, regenerate, and re-classify the result to track the margin (Supp.\ Sec.~S4).

\paragraph{Shortcut discovery (\shortcutflag{} in \cref{fig:teaser}).}
Medical classifiers often learn shortcuts such as acquisition settings~\cite{DeGrave2021}.
Because the axes split every decision exactly, a shortcut's evidence cannot hide from the concept basis, and the audit asks which axes carry it.
We therefore apply the naming test to the acquisition and context variables in the dataset metadata and flag any axis whose strongest effect is such a variable.
A flag only names a suspect, which \cref{sec:audit} then tests.

%% file: sec/4_experiments.tex
\section{Experiments}
\label{sec:setup}

\paragraph{Datasets.}
Five imbalanced tasks span four medical modalities (Supp.\ Sec.~S5.1).
Four datasets are public:
\emph{Skin}, ISIC 2018 Task 3~\cite{HAM10000,ISIC2018} (10,015 dermoscopic images, 7 classes);
\emph{PBC}~\cite{Acevedo2020} (17,092 images, 8 classes);
\emph{BMC}~\cite{Matek2021} (171,374 images, 21 classes), stained blood and bone-marrow smears;
and \emph{Chest}, NIH ChestX-ray14~\cite{ChestXray14} (112,120 images, 14 findings, multi-label).
We use official data splits for the public datasets, when available.
\emph{OBM}, 40,920 stain-free white blood cells (5 classes), is a private, IRB-approved oblique back-illumination microscopy dataset split by image session.
For evaluation only, we use the ISIC 2018 Task 2 dermoscopic-structure masks~\cite{ISIC2018}, the PBC maturation-stage labels, and the acquisition metadata of each dataset, which never enter training.

\paragraph{Implementation.}
DINOv3 ViT-B/16~\cite{DINOv3} is frozen and read at all 12 layers with 12 concepts each ($U=144$, $224\times224$ input).
Per dataset, three bottleneck seeds are trained and their $S=3$ bases learned jointly.
Training takes one RTX A5500 per seed, and inference adds 4.3M parameters on top of the encoder pass.
The number of axes $K$ and the loss weights are tuned per dataset on validation, and all other settings are shared across datasets (training details in Supp.\ Sec.~S1).
The naming VLMs score images against a fixed term list per dataset.

\paragraph{Baselines and metrics.}
We retrain every code-released concept-free baseline on our splits, score it in its final interpretable form, and cite it in its \cref{tab:main} row (protocols in Supp.\ Sec.~S5.2).
All methods are scored by balanced accuracy (macro AUROC for chest) over three seeds.
As domain-specific references, \cref{tab:main} gives the range of the best published results of models trained on the same public datasets and tested on their official test sets, each under its own protocol, including in-domain pretraining, test-time tuning, or higher resolution (sources in the \cref{tab:main} footnote).
\cref{tab:ablation} measures are defined in Supp.\ Sec.~S6.
Attention coverage counts patches inside the half-mass core of any slot, shared errors compares layers at least four apart on test images, axes per decision counts axes carrying 80\% of an image's positive margin (class bias excluded), and seed match correlates the worst-matched axis across seeds.

\providecommand{\sd}[1]{{\scriptsize$\pm$#1}}
\begin{table}[t]
  \caption{\textbf{Classification accuracy on five medical tasks.}
  Mean test balanced accuracy over 3 seeds (macro AUROC for chest), $\pm$\,s.d. over seeds.
  Best in bold, second underlined.
  Shaded rows read several layers.
  $\S$~Softmax averaged over 12 per-layer CLS$\|$mean probes.
  $\ddagger$~Multi-label adaptation for chest (Supp.\ Sec.~S5.2).}
  \label{tab:main}
  \centering
  \resizebox{\columnwidth}{!}{%
  \begin{tabular}{@{}l*{5}{c}@{}}
    \toprule
    \textbf{Method} & \textbf{Skin} & \textbf{PBC} & \textbf{OBM} & \textbf{BMC} & \textbf{Chest} \\
    \midrule
    \multicolumn{6}{@{}l}{\emph{(i) Frozen readout}} \\
    \quad Linear probe, last layer       & 71.1\sd{0.2} & 96.6\sd{0.1} & 62.5\sd{0.3} & 59.4\sd{0.2} & 75.2\sd{0.0} \\
    \rowcolor{rowshade}
    \quad Linear probe, 12-layer vote$^{\S}$ & 76.8\sd{0.1} & 98.3\sd{0.0} & 69.8\sd{0.3} & 69.8\sd{0.3} & 76.8\sd{0.3} \\
    \quad Attentive probe~\cite{AIM} (ICML'24)      & 76.4\sd{0.4} & 98.3\sd{0.1} & 68.5\sd{0.6} & 70.1\sd{1.1} & 79.2\sd{0.2} \\
    \rowcolor{rowshade}
    \quad Head2Toe~\cite{Head2Toe} (ICML'22)            & 75.2\sd{0.5} & 97.6\sd{0.3} & 72.7\sd{0.3} & 65.5\sd{0.7} & 77.4\sd{0.4} \\
    \midrule
    \multicolumn{6}{@{}l}{\emph{(ii) Prototype / compact head}} \\
    \quad PNP~\cite{PNP} (CVPR'25)                 & 73.8\sd{0.3} & 97.9\sd{0.2} & 67.7\sd{0.4} & 71.4\sd{0.4} & 61.0\sd{0.7}$^\ddagger$ \\
    \quad PIP-Net~\cite{PIPNet} (CVPR'23)             & 69.6\sd{0.7} & 97.9\sd{0.1} & 68.9\sd{0.7} & 66.5\sd{0.4} & 75.2\sd{0.3} \\
    \rowcolor{rowshade}
    \quad MCPNet$^{*}$~\cite{MCPNet} (CVPR'24)        & 70.2\sd{0.2} & 98.9\sd{0.2} & 74.8\sd{0.4} & 73.2\sd{0.6} & 61.4\sd{1.3}$^\ddagger$ \\
    \quad ProtoViT$^{*}$~\cite{ProtoViT} (NeurIPS'24)   & 76.6\sd{0.4} & \underline{99.3}\sd{0.2} & \underline{77.7}\sd{0.8} & \underline{76.2}\sd{0.5} & 78.0\sd{0.4} \\
    \midrule
    \multicolumn{6}{@{}l}{\emph{(iii) Text-defined concepts}} \\
    \quad LF-CBM$^{\dagger}$~\cite{LFCBM} (ICLR'23)  & 65.5\sd{0.6} & 93.6\sd{0.1} & 47.0\sd{0.2} & 50.6\sd{1.3} & 79.7\sd{0.5} \\
    \quad PCBM-h$^{\dagger}$~\cite{PCBM} (ICLR'23)  & 77.9\sd{0.4} & 97.6\sd{0.1} & 60.4\sd{0.3} & 67.3\sd{0.6} & \underline{80.2}\sd{0.6} \\
    \quad Explicd$^{\dagger}$~\cite{Explicd} (MICCAI'24)& 76.1\sd{0.2} & 98.4\sd{0.3} & 66.4\sd{0.4} & 74.3\sd{0.3} & 75.6\sd{0.2}$^\ddagger$ \\
    \midrule
    \multicolumn{6}{@{}l}{\emph{(iv) Learned concepts}} \\
    \quad DN-CBM$^{\dagger}$~\cite{DNCBM} (ECCV'24)  & 66.5\sd{0.4} & 96.2\sd{0.1} & 65.2\sd{0.2} & 55.5\sd{0.8} & 79.2\sd{0.3} \\
    \quad BotCL$^{*}$~\cite{BotCL} (CVPR'23)         & 79.5\sd{0.6} & 99.1\sd{0.3} & 77.3\sd{0.7} & 69.2\sd{1.6} & 75.3\sd{0.4} \\
    \cmidrule{1-6}
    \quad Ours (single-layer probe)      & \underline{80.0}\sd{0.2} & 99.0\sd{0.1} & 73.3\sd{0.3} & 74.7\sd{0.4} & \underline{80.2}\sd{0.2} \\
    \rowcolor{rowshade}
    \quad \textbf{Ours}                  & \textbf{86.1}\sd{0.2} & \textbf{99.4}\sd{0.0} & \textbf{78.6}\sd{0.3} & \textbf{80.6}\sd{0.3} & \textbf{81.3}\sd{0.0} \\
    \midrule
    \emph{Domain SOTA} & {\small 85.3--87.5}$^{a}$ & {\small 98.9--99.0}$^{b}$ & --- & {\small 71.4--75.7}$^{c}$ & {\small 82.3--84.4}$^{d}$ \\
    \bottomrule
  \end{tabular}}
  \par\vspace{3pt}
  {\scriptsize\raggedright
  Encoders are frozen DINOv3-B unless marked: $^{\dagger}$~frozen MedSigLIP; $^{*}$~encoder fine-tuned, as in the original paper (MCPNet: ResNet-50).
  Literature ranges, each under its own protocol:
  $^{a}$~\cite{Shen2022,Yao2022};
  $^{b}$~\cite{Firat2024,MedMNISTPlus};
  $^{c}$~\cite{Matek2021,DinoBloom,CytoDINO};
  $^{d}$~\cite{Xiao2023,A3Net,ArkPlus}.\par}
\end{table}

%% file: sec/5_results.tex
\section{Results}
\label{sec:results}

\subsection{Comparable accuracy to SOTA from concept bottleneck}
\label{sec:acc}

Our model is the most accurate frozen-encoder model on all five datasets, and its basis maintains this accuracy exactly.
It matches the domain-specific SOTA on skin and PBC and exceeds it on BMC.
On chest, SOTA models lead by 1.0--3.1 points with X-ray pretraining or higher input resolution.
Three DDxBasis design choices give the gain over the linear probe: a strong per-layer bottleneck, reading every layer, and a merge that does not overfit.

\paragraph{A single layer of our bottleneck already beats single-layer models.}
Existing prototype and concept models read the last layer only and classify close to the attentive probe.
Frozen prototype models (PNP, PIP-Net) trade accuracy for interpretability, scoring below the attentive probe in 8 of their 10 results.
Our best single layer exceeds the best frozen single-layer baseline on the DINOv3-B encoder on skin (+3.6, attentive probe), BMC (+3.3, PNP), and chest (+1.0, attentive probe), due to our attention design (\cref{tab:ablation}a).
Attention pooling raises per-layer accuracy from 67.1 (mean pooling) to 73.8--76.4.
Competing slots cover the whole image, while MQA covers only a quarter, and with covariance pooling they add 2.6 points over MQA and give sharper, stabler image sectioning across seeds (overlap 0.17 \vs 0.22, seed agreement 0.66 \vs 0.48).

\paragraph{Layer sum outperforms learned merges (\cref{tab:ablation}b).}
The plain sum of the 12 layers consistently lifts our model above its best layer (by 6.1 points on skin, 5.3 on OBM, 5.9 on BMC, and 1.1 on chest, \cref{tab:main}).
The same vote over 12 per-layer linear probes reaches only 76.8 on skin (\cref{tab:main}), so the gain needs our per-layer bottleneck, not depth ensembling alone.
The best layer lies mid-depth (layers 6--9), yet no layer holds all cues.
On skin, layers four or more apart share only 37\% of their errors (\cref{tab:ablation}a), and at least one layer is correct on 95.9\% of test images.
The gates make the layers more complementary, so the summed model leads (86.1 \vs 84.3) although each gated layer is 1.4 points weaker.
Accuracy on rare or easily confused classes improves the most, such as basal cell carcinoma (+7.9) and melanoma (+6.2) on skin and the rarer half of the BMC classes (+8.1 \vs +3.9 for the common half).
Learned merges fit the noise of small medical training sets.
Fusing the features of all layers with a head always falls below the plain sum.
Head2Toe~\cite{Head2Toe}, ScalarMix~\cite{ELMo}, VFusion~\cite{VFusion}, and ALF~\cite{ALF} fall 7.5 to 12.6 points below our model (Supp.\ Sec.~S5.2).
Layer weights or a greedy layer subset fitted on validation at best tie it.
The plain sum (\cref{eq:sum}) is thus the most cost-effective merge and keeps each layer's share of a logit explicit for exact explanation (\cref{sec:basisres}).

\begin{table}[t]
  \caption{\textbf{Ablations.}
  (a) Per-layer pooling on skin, mean accuracy over the 12 layers.
  (b) Merging across depth, where feature fusion shows the best of six cross-layer readouts per task.
  (c) The final concept basis per task (validation).
  Metrics in \cref{sec:setup}, best in bold, second underlined, full versions in Supp.\ Tables S7, S6, and S5.}
  \label{tab:ablation}
  \centering
  \resizebox{\columnwidth}{!}{%
  \begin{tabular}{@{}l*{5}{c}@{}}
    \toprule
    \textbf{\emph{(a) Within a layer (skin)}}
      & \makecell{\textbf{Mean}\\\textbf{acc.}\,$\uparrow$}
      & \makecell{\textbf{Att. coverage}\,$\uparrow$\\\textbf{(overlap}\,$\downarrow$\textbf{)}}
      & \makecell{\textbf{Slot}\\\textbf{similarity}\,$\downarrow$}
      & \makecell{\textbf{Shared}\\\textbf{errors}\,$\downarrow$}
      & \makecell{\textbf{Seed}\\\textbf{agreement}\,$\uparrow$} \\
    \midrule
    Probe heads (CLS$\|$mean)   & 67.1 & \textbf{1.00} (--)   & --   & 0.43 & -- \\
    MQA ($n{=}12$)              & 73.8 & 0.26 (0.58) & 0.50 & \underline{0.39} & 0.01 \\
    Slot attention (SA, $n{=}12$) & 74.5 & \textbf{1.00} (0.22) & \textbf{0.45} & \underline{0.39} & 0.48 \\
    SA + covariance             & \textbf{76.4} & \underline{0.99} (\underline{0.17}) & 0.59 & 0.40 & \textbf{0.66} \\
    \textbf{+ gates (ours)}     & \underline{75.0} & \underline{0.99} (\textbf{0.16}) & \underline{0.48} & \textbf{0.37} & \underline{0.60} \\
    \midrule
    \textbf{\emph{(b) Merging across depth}} & \textbf{Skin} & \textbf{PBC} & \textbf{OBM} & \textbf{BMC} & \textbf{Chest} \\
    \midrule
    Feature fusion, one head    & 75.2 & 98.3 & 71.3 & 65.1 & 77.4 \\
    Greedy layer subset         & 84.6 & 99.2 & 77.4 & 79.4 & \textbf{81.5} \\
    Learned layer weights       & 85.6 & 99.2 & 77.7 & 79.4 & 80.8 \\
    \textbf{Plain logit sum (ours)} & \textbf{86.1} & \textbf{99.4} & \textbf{78.6} & \textbf{80.6} & 81.3 \\
    \midrule
    \textbf{\emph{(c) Concept basis}} & \textbf{Skin} & \textbf{PBC} & \textbf{OBM} & \textbf{BMC} & \textbf{Chest} \\
    \midrule
    Effective axes\,$\uparrow$     & 5.42/6 & 3.95/4 & 3.88/4 & 5.62/6 & 5.61/6 \\
    Axes per decision\,$\downarrow$ & 2.0 (26) & 1.7 (21) & 1.0 (20) & 2.7 (27) & 1.0 (15) \\
    Seed match\,$\uparrow$         & .89 & .99 & .98 & .85 & .88 \\
    Stability\,$\uparrow$          & .82 (.72) & .96 (.86) & .78 (.60) & .97 (.92) & .93 (.81) \\
    \bottomrule
  \end{tabular}}
\end{table}

\paragraph{Fine-tuned and residual baselines come closest.}
We train a baseline end to end with its published protocol when its frozen-encoder adaptation falls below the linear probe.
On OBM, whose grayscale phase contrast lies furthest from natural images, ProtoViT and BotCL reach 77.3--77.7 (\vs 78.6) by fine-tuning all 85.7M encoder weights per task.
PCBM-h~\cite{PCBM} comes within 1.1 points on chest with an unexplained residual that carries 35--79\% of its margin.
LF-CBM~\cite{LFCBM} falls below the linear probe on BMC, where MedSigLIP reads only 4 of 30 morphology terms.

\subsection{Exact, compact, distinct, and reproducible basis}
\label{sec:basisres}

A basis of four to six axes splits the stage-1 decisions exactly, and nearly all of them are decorrelated (\cref{tab:ablation}c, effective axes).
A decision typically rests on one to three axes that carry 80\% of its positive margin, where the bottleneck needs 15 to 27 concepts.
However, different classes are most often led by different axes, three to five distinct axes per dataset.
The most used axis carries on average only 33\% of the margin (31--35\% across datasets), so different decisions draw on different subsets of the basis and together need all of it.
Within a class, the position tracks the axis margin between its two pole classes (Spearman $\rho$ 0.85 on average, 0.63--0.99 per dataset, Supp.\ Sec.~S9.1), so where a case lies on an axis reflects its evidence.
The jointly trained bases are reproducible (seed match 0.85 to 0.99) and more stable under augmentation than the concepts they regroup, by 0.11 on average.
Without the consensus term the median axis still matches (0.84) but the worst does not (0.19, Supp.\ Table~S10), so agreement on every axis is enforced, not emergent.

\begin{figure}[t]
  \centering
  \includegraphics[width=\columnwidth]{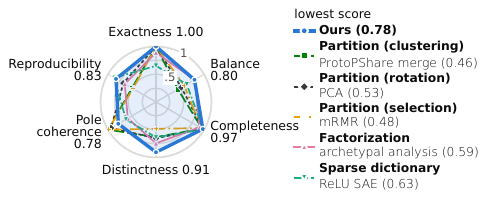}
  \caption{\textbf{Structural comparison of concept bases.}
  Ours vs.\ the best of five families on six properties (skin, 0--1, outward is better).
  Definitions and all methods in Supp.\ Sec.~S8.}
  \label{fig:radar}
\end{figure}

\paragraph{Each objective term delivers its target property (\cref{fig:radar}).}
We compare our basis with 19 condensations of the same 144 concepts into six dimensions.
We evaluate these properties to verify the training objective before testing clinical meaning.
Partitions of the concepts (ProtoPShare merge, PCA, and mRMR) reach near-perfect pole coherence by construction but let one axis carry up to half of each decision (balance 0.46--0.59).
Ours trades some pole coherence (0.78) for balance (0.80) and distinctness (0.91).
Sparse dictionaries and factorizations leave part of the margin unexplained (exactness 0.66, 0.89).
Our basis is exact like a partition and balanced like a factorization, and each objective term is needed for one of these properties (Supp.\ Table~S10).

\subsection{The concept basis reads as clinical differentials}
\label{sec:clinical}

\begin{figure*}[t]
  \centering
  \includegraphics[width=\textwidth]{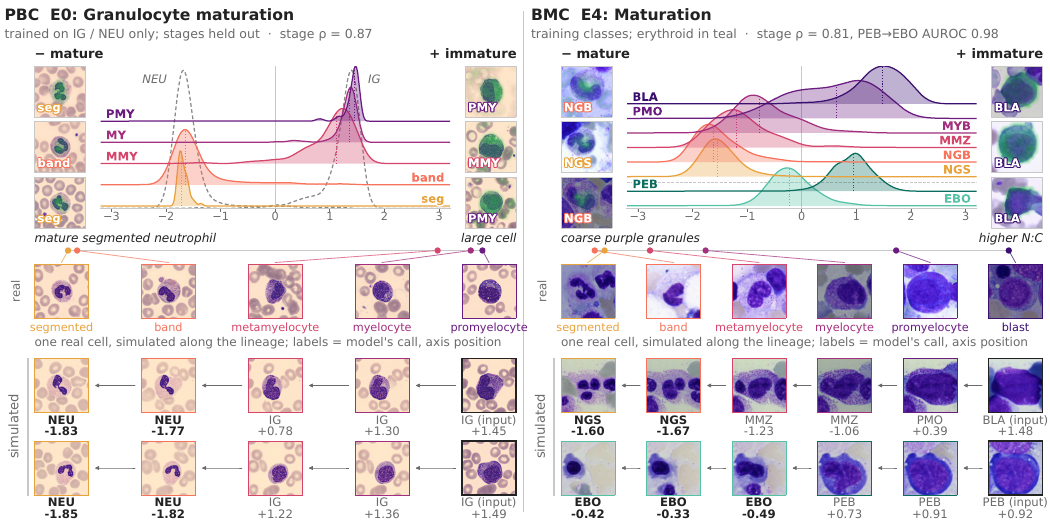}
  \caption{\textbf{Maturation axes learned on PBC (left) and BMC (right) without ordered stage labels.}
  Each axis orders cells from immature (+) to mature ($-$) (Spearman $\rho=0.87$ and $0.81$).
  Shown per dataset: pole atlases with pole maps, stage densities along the axis, one real cell sample per stage (\emph{real}), and counterfactual edits along the maturation axis (\emph{simulated}).}
  \label{fig:maturation}
\end{figure*}

\paragraph{The axes discover differential clinical reasoning beyond independent classes (\cref{fig:maturation}).}
On PBC, the E0 axis orders the five granulocyte maturation stages on held-out data (stage $\rho$ 0.87), although the classifier sees only two merged labels, immature granulocyte (IG) and neutrophil (NEU).
Within the IG class alone, E0 still correctly orders PMY $<$ MY $<$ MMY (stage $\rho$ 0.64), whereas the classifier's own IG-or-NEU call carries no such order (0.06).
On BMC, where each stage is an unordered training class, the E4 axis orders several cell lineages on test images, placing blasts through segmented neutrophils in maturation order (stage $\rho$ 0.81) and separating proerythroblasts from erythroblasts (AUROC 0.98).
Edited along this lineage with the flow U-Net, a real blast gains granules, shrinks, and its nucleus indents and finally segments.
These counterfactuals move mainly the edited axis, 4.1--14.3 times more than the others, and an independently trained ConvNeXt-T~\cite{ConvNeXt} sees edited PBC cells mature as much as real cells (Supp.\ Sec.~S9.2).
In both datasets the ordering is weaker on the other axes (average stage $\rho$ 0.31 and 0.33).
All named axes are in Supp.\ Table~S4.

\paragraph{The axes expose rare subtypes hidden inside a training label.}
Clinically important subtypes are often too rare to be labeled as separate classes and are merged into a larger one.
This hidden stratification can conceal poor performance on the subtype~\cite{OakdenRayner2020}.
On BMC, hairy cells ($n=409$) are outnumbered 64 to 1 by the similar lymphocytes ($n=26{,}242$).
With hairy cells merged into lymphocytes during training, a single axis (E3) still separates them at test AUROC 0.81, against 0.86 when they are trained with separate labels.
The same holds for immature lymphocytes ($n=65$, AUROC 0.89 \vs 0.98) and for faggot cells against promyelocytes ($n=47$ \vs 11,994, 0.92 \vs 0.91, Supp.\ Sec.~S9.3).
A basis learned from coarse diagnostic labels can flag rare, clinically decisive subtypes for expert review without their own labels.

\paragraph{The skin basis is validated by dermatologists' annotations.}
Five of the six skin axes (Supp.\ Sec.~S9.4) are contrasts that dermatologists draw, such as melanoma against its keratotic mimics, nevi against other lesions, and keratinocyte carcinomas against benign keratosis.
The sixth tracks acquisition, a potential shortcut tested in \cref{sec:audit}.
As a reference unseen in training, we use the ISIC 2018 Task 2 masks~\cite{ISIC2018}, in which dermatologists marked five structures (pigment network, negative network, streaks, globules, and milia-like cysts) on 2,590 lesions.
Our basis agrees with them in three ways.
Four of the five checkable axis names match the marked structures.
Within a class, the melanocytic axis E1 separates globules, pigment network, and streaks from milia-like cysts, which mirrors the first step of the common two-step dermoscopy algorithm~\cite{Marghoob2010}.
Finally, the six axis positions alone predict the structures better than the diagnosis label (AUROC 0.69 \vs 0.58).
Without any annotation, the skin basis follows the structures that dermatologists use to tell these lesions apart, so its explanations can be checked against what a dermatologist sees.

\subsection{The concept basis makes the model auditable}
\label{sec:audit}

\paragraph{Axis statistics support auditing borderline errors.}
PBC errors concentrate on the two transitional stage boundaries along the maturation axis (E0), where hematologists also disagree, with 65 of 70 band-form neutrophils called immature granulocytes and 23 of 26 metamyelocytes called neutrophils.
In all 96 errors, E0 carries most of the error margin (58\% on average, Supp.\ Sec.~S9.5).
On skin, axis E1 (melanocytic \vs keratinocytic) decides 133 of 134 nevi called melanoma, so a reader can flag the failed differential for borderline-case review.
As an oracle upper bound~\cite{CBM}, correcting the most misleading axis per image, by replacing its evidence with the true class's mean evidence, raises skin validation balanced accuracy from 82.3 to 94.8, as much as correcting 12 to 24 concepts (Supp.\ Table~S13).

\paragraph{A planted shortcut lands on one axis and is removed with it.}
Following \citet{Adebayo2022}, we draw a green dot on skin training images at two strengths (90\% of melanomas and 5\% of others, or 80\% and 10\%).
The model learns both, losing 25.1 and 11.5 points on reversed test images (dot moved onto other classes), while a dot-free model is unaffected (84.6 to 84.9, Supp.\ Table~S14).
The 144 concepts spread the dot over 12 seed-specific units, whereas one pole of the basis carries most of the dot's melanoma evidence (83\% and 74\%), and MedSigLIP names that pole ``a green dot'', ahead of clinical terms.
Removing that axis raises reversed accuracy by 19.9 points (58.1 to 78.0) at the strong setting and by 5.2 points (75.3 to 80.5) at the weak one, whereas removing the best other axis costs 1.2 and 0.6 points on average.
The removal also raises accuracy on dot-free images by 1.6 and 1.2 points.
The basis gathers the scattered shortcut into one named axis that is easy to find and remove.

\paragraph{The axes discover potential dataset shortcuts, and a test decides whether the model is affected.}
Although the basis never sees acquisition variables, five axes are flagged because an acquisition or context variable is their strongest held-out effect (\cref{sec:reading}).
On chest, axis E2 tracks view position (PA/AP AUROC 0.86), which influences the prediction (test AUROC 0.88 when view agrees with the class prior, 0.66 against it).
View position is a known confound in chest radiography, because AP films are taken at the bedside of sicker, often supine patients and magnify the heart~\cite{Zech2018}.
Within PA films, E2 is renamed hyperinflation.
The other suspects are cleared: the classifiers do not rely on film rotation (chest E4), crowding (BMC E2), or the skin acquisition axis (E0) (Supp.\ Sec.~S9.7).
The basis turns a vague concern about acquisition bias into named suspects and tests which affect the diagnosis.

\subsection{Limitations}
\label{sec:limits}

Clinical names are mined post hoc with a fixed dictionary and VLMs, so not every axis maps to a known term.
Exactness makes the axes faithful to the model, but no axis is guaranteed to be one clinical variable.
Names and counterfactual edits are not yet checked by blinded clinicians.
Hyperparameters such as $K$ are tuned per task, not exhaustively, and the 224\,px input may miss fine chest details separating nodules from masses.
Apart from the planted one, the audit diagnoses existing shortcuts such as view position but leaves fixing them, by curating data and retraining, to future work.

%% file: sec/6_conclusion.tex
\section{Conclusion}
\label{sec:conclusion}

Our concept bottleneck, reading every layer of a frozen encoder, is the most accurate frozen-encoder model, reaching near-SOTA classification across tasks.
To read a differential diagnosis from this strong classifier, the concept basis condenses concepts exactly into reproducible, well-structured two-pole axes, each a distinct, namable differential between groups of diagnoses, that recover cell lineages and rare subtypes, agree with dermatologists' annotations, and expose planted and real shortcuts without concept labels.

%% file: sec/X_suppl_arxiv.tex
\clearpage
\twocolumn[{\centering\Large\textbf{\thetitle}\\\vspace{0.5em}Supplementary Material\\\vspace{1.0em}
\raggedright\normalsize
\vspace{0.8em}{\large\bfseries Contents}\par\vspace{0.3em}
{\small\begin{tabularx}{\textwidth}{@{}l>{\raggedright\arraybackslash}X@{}}
\toprule
Section & Title \\
\midrule
\hyperref[sec:S1]{\textbf{S1}} & \hyperref[sec:S1]{\textbf{Implementation details}} \\
\hyperref[sec:S2]{\textbf{S2}} & \hyperref[sec:S2]{\textbf{Naming poles: scores, admissible classes and checks}} \\
\hyperref[sec:S3]{\textbf{S3}} & \hyperref[sec:S3]{\textbf{Content codes and descriptor residual}} \\
\hyperref[sec:S4]{\textbf{S4}} & \hyperref[sec:S4]{\textbf{Generator details}} \\
\hyperref[sec:S5]{\textbf{S5}} & \hyperref[sec:S5]{\textbf{Datasets and accuracy}} \\
\hspace*{1.2em}\hyperref[sec:S5.1]{S5.1} & \hspace*{1.2em}\hyperref[sec:S5.1]{Datasets} \\
\hspace*{1.2em}\hyperref[sec:S5.2]{S5.2} & \hspace*{1.2em}\hyperref[sec:S5.2]{Accuracy} \\
\hyperref[sec:S6]{\textbf{S6}} & \hyperref[sec:S6]{\textbf{Ablations of pipeline designs on all five datasets}} \\
\hspace*{1.2em}\hyperref[sec:S6.1]{S6.1} & \hspace*{1.2em}\hyperref[sec:S6.1]{Within a layer (main Table 2a)} \\
\hspace*{1.2em}\hyperref[sec:S6.2]{S6.2} & \hspace*{1.2em}\hyperref[sec:S6.2]{Across depth (main Table 2b)} \\
\hspace*{1.2em}\hyperref[sec:S6.3]{S6.3} & \hspace*{1.2em}\hyperref[sec:S6.3]{Concept basis (main Table 2c)} \\
\hyperref[sec:S7]{\textbf{S7}} & \hyperref[sec:S7]{\textbf{Generalization to other foundation-model backbones}} \\
\hyperref[sec:S8]{\textbf{S8}} & \hyperref[sec:S8]{\textbf{Structural evaluation of the concept basis}} \\
\hyperref[sec:S9]{\textbf{S9}} & \hyperref[sec:S9]{\textbf{Extended results}} \\
\hspace*{1.2em}\hyperref[sec:S9.1]{S9.1} & \hspace*{1.2em}\hyperref[sec:S9.1]{Position–margin agreement} \\
\hspace*{1.2em}\hyperref[sec:S9.2]{S9.2} & \hspace*{1.2em}\hyperref[sec:S9.2]{Maturation axes and counterfactual edits} \\
\hspace*{1.2em}\hyperref[sec:S9.3]{S9.3} & \hspace*{1.2em}\hyperref[sec:S9.3]{Axis-based differentials for rare abnormal cells, with and without their labels} \\
\hspace*{1.2em}\hyperref[sec:S9.4]{S9.4} & \hspace*{1.2em}\hyperref[sec:S9.4]{Skin axes against dermatologists' annotations} \\
\hspace*{1.2em}\hyperref[sec:S9.5]{S9.5} & \hspace*{1.2em}\hyperref[sec:S9.5]{Errors and oracle interventions} \\
\hspace*{1.2em}\hyperref[sec:S9.6]{S9.6} & \hspace*{1.2em}\hyperref[sec:S9.6]{Planted shortcut} \\
\hspace*{1.2em}\hyperref[sec:S9.7]{S9.7} & \hspace*{1.2em}\hyperref[sec:S9.7]{Acquisition shortcuts} \\
 & \hyperref[sec:refs]{\textbf{References}} \\
\bottomrule
\end{tabularx}}\par
\vspace{0.8em}{\large\bfseries List of tables}\par\vspace{0.3em}
{\small\begin{tabularx}{\textwidth}{@{}l>{\raggedright\arraybackslash}Xl@{}}
\toprule
Table & Title & Section \\
\midrule
\hyperref[tab:S1]{Table S1} & \hyperref[tab:S1]{Term lists and admissible terms} & S2 \\
\hyperref[tab:S2]{Table S2} & \hyperref[tab:S2]{Differential and pole names of each axis} & S2 \\
\hyperref[tab:S3]{Table S3} & \hyperref[tab:S3]{Names checked against annotations} & S2 \\
\hyperref[tab:S3a]{Table S3a} & \hyperref[tab:S3a]{The six skin axis positions predict every structure better than the diagnosis} & S9.4 \\
\hyperref[tab:S4]{Table S4} & \hyperref[tab:S4]{Axis labels} & S2 \\
\hyperref[tab:S5a]{Table S5a} & \hyperref[tab:S5a]{Training settings of the Table 1 baselines} & S5.2 \\
\hyperref[tab:S5]{Table S5} & \hyperref[tab:S5]{Main Table 2a on all five datasets} & S6.1 \\
\hyperref[tab:S6]{Table S6} & \hyperref[tab:S6]{Every readout across depth (full main Table 2b)} & S6.2 \\
\hyperref[tab:S7]{Table S7} & \hyperref[tab:S7]{The basis on all five datasets (full main Table 2c)} & S6.3 \\
\hyperref[tab:S8]{Table S8} & \hyperref[tab:S8]{The logit sum with other frozen backbones} & S7 \\
\hyperref[tab:S9]{Table S9} & \hyperref[tab:S9]{All 19 condensations of main Fig. 3} & S8 \\
\hyperref[tab:S10]{Table S10} & \hyperref[tab:S10]{Ablation of the basis terms} & S8 \\
\hyperref[tab:S11]{Table S11} & \hyperref[tab:S11]{Counterfactual edits move the requested axis and keep the cell} & S9.2 \\
\hyperref[tab:S12]{Table S12} & \hyperref[tab:S12]{Abnormal cells against their parent class on a training-selected axis} & S9.3 \\
\hyperref[tab:S13]{Table S13} & \hyperref[tab:S13]{One corrected axis does the work of 12 to 24 corrected concepts} & S9.5 \\
\hyperref[tab:S14]{Table S14} & \hyperref[tab:S14]{A planted dot lands on one axis at both strengths and is removed with it} & S9.6 \\
\hyperref[tab:S15]{Table S15} & \hyperref[tab:S15]{Acquisition audit} & S9.7 \\
\bottomrule
\end{tabularx}}\par
\vspace{0.8em}{\large\bfseries List of figures}\par\vspace{0.3em}
{\small\begin{tabularx}{\textwidth}{@{}l>{\raggedright\arraybackslash}Xl@{}}
\toprule
Figure & Title & Section \\
\midrule
\hyperref[fig:S1]{Figure S1} & \hyperref[fig:S1]{Axis atlases of the four public datasets} & S2 \\
\hyperref[fig:S2]{Figure S2} & \hyperref[fig:S2]{Per-layer accuracy and the cross-layer evidence sum} & S6.2 \\
\bottomrule
\end{tabularx}}\par
}]
\clearpage

\setcounter{section}{0}
\renewcommand{\thesection}{S\arabic{section}}
\renewcommand{\thesubsection}{\thesection.\arabic{subsection}}

\section{Implementation details}\label{sec:S1}

\paragraph{Backbone and concept bottleneck.} The encoder $f$ of main Sec. 3.1 is DINOv3 ViT-B/16 \cite{DINOv3} ($L=12$, $D=768$, patch size 16, four register tokens). We use $224\times224$ inputs ($N=196$ patches) and read all 12 layers with $M=12$ concepts (slots) each ($U=144$, $d=128$). The concept bottleneck is trained in two phases per dataset and seed, both with AdamW (lr $4\cdot10^{-4}$, batch 32) and a reconstruction weight of 3. The first phase trains the slots and per-layer heads without presences for 30 epochs. The second phase adds the presences (learning rate $10^{-2}$ for the presence head) and fine-tunes from the first for another 30 epochs. The non-negative class weights $W$ carry an $\ell_1$ penalty of 0.01, a hinge of weight 1 keeps each presence active on 10–90\% of a batch, and the class-entropy hinge of main Sec. 3.2 has weight 0.1. The multi-label chest loss weights the positives of each finding by $\sqrt{\text{negatives}/\text{positives}}$, clipped to $[1,10]$. The number of presences per concept follows the size of the label space. We use one presence ($G=1$) per concept for the label spaces of up to eight classes (skin, PBC, and OBM). The two largest label spaces, BMC ($C=21$) and chest ($C=14$), use $G=3$ presences per concept. Each presence then adds evidence through its own non-negative weights, and wherever the main text uses one presence per concept (positions, pole maps, pole coherence, content codes) the strongest one stands in, $\bar g_u(x)=\max_\gamma g_{u\gamma}(x)$, which equals $g_u(x)$ when $G=1$. All sums run over (concept, presence) units, so logits and axes stay exact.

\paragraph{Concept basis.} The basis is trained for 150 epochs, with AdamW (lr $10^{-3}$, one-cycle schedule) and batch 256. It uses $S=3$ jointly learned bases, one per concept-bottleneck seed. The basis weights are $\lambda_{\text{bal}}=10$ with $[\kappa_{\min},\kappa_{\max}]=[0.08,0.35]$ and consensus weights of 3 on positions ($\lambda_{\text{agr}}$, Eq. \hyperref[eq:S1.4]{S1.4}), 2 on content codes and 1 on class evidence. The temperature $\tau$ of main Eq. (4) is annealed from 1 to $\tau_{\min}=0.02$, and the factor initialization uses $\lambda_0=0.5$. Content codes have dimension 16, the descriptor residual keeps $\eta^2\le0.05$, and $\lambda_{\text{coh}}=\lambda_{\text{red}}=3$. These values are shared by all datasets. The number of axes and three loss weights, $\lambda_{\text{pole}}\in\{1,3\}$, $\lambda_{\text{vol}}\in\{3,5\}$, and $\lambda_{\text{cmp}}\in\{0,0.3\}$, are chosen per dataset on validation (Sec. \hyperref[sec:S6]{S6}), with $K=6$ for skin, BMC, and chest and $K=4$ for PBC and OBM.

\paragraph{Compute and inference.} Each seed of the concept bottleneck and each set of $S$ jointly learned bases is trained on a single NVIDIA RTX A5500 GPU (24 GB). At inference, one pass of the frozen encoder (85.7M parameters) feeds the 12 per-layer pools and heads (4.3M parameters in total), which return the prediction together with its exact decomposition into layers, concepts and axes. The reconstruction decoders (9M parameters) are used only in training, and the generator (an 8.3M-parameter U-Net and 0.9M for its condition tokens) only for counterfactual edits on request (Sec. \hyperref[sec:S4]{S4}).

\paragraph{Basis losses (main Sec. 3.3).} The four terms of main Sec. 3.3 make the axes distinct, balanced, coherent, and reproducible. Each acts on within-class quantities, where $\mathrm{wc}(\cdot)$ removes the class mean and standardizes over the batch.

\textit{Distinct axes.} A log-determinant volume \cite{Kulesza2012} on $\hat\Sigma$, the $K\times K$ correlation matrix of the within-class positions $\tilde p=\mathrm{wc}(p)$, penalizes collinear axes,
\begin{equation*}
\mathcal{L}_{\text{vol}}=-\frac{1}{K}\log\det\big(\hat\Sigma+\epsilon I\big). \tag{S1.1}\label{eq:S1.1}
\end{equation*}

\textit{Balanced axes.} A load-balancing band keeps the average margin share $\bar\rho_j$ of every axis between $\kappa_{\min}$ and $\kappa_{\max}$, where $\rho_j(x)=[\mu_j(x)]_+/\sum_{j'}[\mu_{j'}(x)]_+$ is the share of axis $j$ in the margin of image $x$,
\begin{equation*}
\begin{aligned}
\mathcal{L}_{\text{bal}}&=\sum_j\big[\bar\rho_j-\kappa_{\max}\big]_+ + \big[\kappa_{\min}-\bar\rho_j\big]_+,\\
\bar\rho_j&=\mathbb{E}_x\,\rho_j(x).
\end{aligned}\tag{S1.2}\label{eq:S1.2}
\end{equation*}

\textit{Coherent poles.} A normalized pole score acts on $R=\mathrm{corr}(\mathrm{wc}(\bar g))$ with a zero diagonal. With $\sigma_j=\pi^{+}_{\cdot j}-\pi^{-}_{\cdot j}$ and $\omega_j=\pi^{+}_{\cdot j}+\pi^{-}_{\cdot j}$, it reads
\begin{equation*}
\mathcal{L}_{\text{pole}}=-\frac{1}{K}\sum_j\frac{\sigma_j^{\top}R\,\sigma_j}{\omega_j^{\top}|R|\,\omega_j}\in[-1,1] \tag{S1.3}\label{eq:S1.3}
\end{equation*}
It reaches $-1$ when every axis is fully coherent.

\textit{Reproducible axes.} The bases of $S$ bottlenecks are learned jointly. All copies start from one PCA–varimax solution of $\mathrm{wc}(\bar g)$ \cite{Kaiser1958}, aligned by the Hungarian algorithm, and a consensus term ties matching axes,
\begin{equation*}
\mathcal{L}_{\text{agr}}=\binom{S}{2}^{-1}\sum_{s<s'}\frac1K\sum_j\Big(1-\mathrm{corr}\big(p^{(s)}_j,p^{(s')}_j\big)\Big), \tag{S1.4}\label{eq:S1.4}
\end{equation*}
with the same consensus on content codes and class evidence.

\textit{Optional compactness.} The normalized entropy of the per-image margin shares, $\mathcal{L}_{\text{cmp}}=-\mathbb{E}_x\sum_j\rho_j(x)\log\rho_j(x)/\log K$, lets each decision rest on few axes. It is optional and not one of the four terms of main Sec. 3.3. Its weight is chosen on validation together with $\lambda_{\text{pole}}$ and $\lambda_{\text{vol}}$. Sec. \hyperref[sec:S3]{S3} gives the full objective.

\clearpage
\section{Naming poles: scores, admissible classes and checks}\label{sec:S2}

\paragraph{Attribute scores.} For each dataset we fix a vocabulary \ensuremath{\mathcal{V}} of clinical attribute terms from the domain's consensus terminology \cite{Kittler2016,Palmer2015,Hansell2008}, listed in Table \hyperref[tab:S1]{S1}. Every image is scored once, as a whole, by a frozen vision–language model, the standard reading used to name units and concepts \cite{CLIPDissect,LFCBM,DNCBM}. For MONET we use its concept-presence score, a softmax over a term prompt and a reference prompt averaged over three templates \cite{MONET}. For MedSigLIP we use the difference of cosine similarities to the prompts ``an image ... with \textit{t}'' and ``an image ... without \textit{t}'' \cite{MedGemma}. Scores are standardized per term on the training split.

\paragraph{Admissible classes.} A term may name a pole only in the classes where the VLM has been shown to read it, with an AUROC of at least 0.7, the conventional bar for acceptable discrimination \cite{Hosmer2013}. In a class with sub-class annotations, the term must separate them within the class. These are the ISIC 2018 Task 2 attribute masks on skin (2,590 images diagnosed as nevus, melanoma, or seborrheic keratosis) and the maturation stages on PBC (band and segmented neutrophils, promyelocytes, myelocytes, and metamyelocytes). In every other class, the term must separate the classes where its finding is expected from the rest. This test uses the annotations to check the VLM, not our model. Naming runs after the model is trained, and the annotations never enter the concept bottleneck or the basis (main Sec. 4). A term that fails the test is never used, so where the VLM reads few terms, as on BMC, the names come from measurements and stage order instead (Table \hyperref[tab:S4]{S4}).

\begin{table*}[t]
\centering
\phantomsection\label{tab:S1}\caption*{\textbf{Table S1. Term lists and admissible terms.} The attribute terms scored by each naming VLM, split into admissible terms and the others, highest first.}
\scriptsize
\setlength{\tabcolsep}{3pt}
\begin{tabularx}{\textwidth}{ll>{\raggedright\arraybackslash}X>{\raggedright\arraybackslash}X}
\toprule
Dataset & VLM & Admissible terms (AUROC) & Other terms (highest AUROC) \\
\midrule
Skin & MONET & red-blue lacunae (0.96); pigment network (0.81); ulceration (0.79); comedo-like openings (0.74); blue-gray ovoid nests (0.74) & central white patch (0.69); scale and keratin (0.69); irregular blotches (0.65); multiple colors (0.64); arborizing vessels (0.64); milia-like cysts (0.63); streaks (0.63); dots and globules (0.63); blue-white veil (0.61); regression structures (0.60); negative network (0.58); asymmetric shape (0.55); shiny white structures (0.51); atypical pigment network (0.50); leaf-like areas (0.47); sharply demarcated border (0.42); strawberry pattern (0.39) \\
PBC & MedSigLIP & orange eosinophilic granules (0.88); gray-blue cytoplasm (0.80); platelet clumps (0.78); large cell (0.77); mature segmented neutrophil (0.74); band-shaped nucleus (0.72); dense dark round nucleus (0.72); indented kidney-shaped nucleus (0.72) & abundant cytoplasm (0.69); high nucleus to cytoplasm ratio (0.69); dark purple basophilic granules (0.65); small cell (0.64); cytoplasmic vacuoles (0.62); small anucleate cell fragments (0.61); round nucleus (0.59); pink granular cytoplasm (0.57); segmented multilobed nucleus (0.56); condensed clumped chromatin (0.54); fine lilac granules (0.50); fine open chromatin (0.37); scant cytoplasm (0.37) \\
BMC & MedSigLIP & smudged ruptured cell (0.81); orange eosinophilic granules (0.78); coarse purple granules (0.70); dark basophilic granules (0.70) & band-shaped nucleus (0.69); prominent nucleoli (0.65); abundant cytoplasm (0.64); pale gray cytoplasm (0.63); large cell (0.62); hairy cytoplasmic projections (0.62); indented kidney-shaped nucleus (0.61); fine azurophilic granules (0.60); Auer rods (0.59); high nucleus to cytoplasm ratio (0.56); deeply basophilic blue cytoplasm (0.55); fine open chromatin (0.55); perinuclear halo (0.55); small cell (0.54); segmented multilobed nucleus (0.54); mature cell (0.51); round nucleus (0.50); cytoplasmic vacuoles (0.49); bare nucleus without cytoplasm (0.47); pink cytoplasm (0.46); cell debris (0.46); eccentric nucleus (0.45); immature blast cell (0.41); agranular cytoplasm (0.39); condensed clumped chromatin (0.38); scant cytoplasm (0.34) \\
Chest & MedSigLIP & perihilar haziness (0.86); blunting of the costophrenic angle (0.81); basal opacity with a meniscus sign (0.81); hyperinflated lungs with flattened diaphragms (0.78); airspace consolidation (0.77); enlarged cardiac silhouette (0.74); visible pleural line without lung markings (0.74); clear lungs (0.73); lung opacity (0.72) & patchy infiltrates (0.64) \\
\bottomrule
\end{tabularx}
\end{table*}

\paragraph{Term contrast (main Sec. 3.4).} For axis \textit{j}, class \textit{c}, the within-class pole atlases $\mathcal{A}^{\pm}_{jc}$ hold the 10\% of training images of \textit{c} (at least three) furthest toward each pole. With \textit{a}ₜ(\textit{x}) the standardized VLM score of term \textit{t} \cite{DNCBM}, term \textit{t} scores
\begin{equation*}
\begin{aligned}
d_{jt}&=\sum_{c\in\mathcal C(t)} w_c\big(\bar a^{+}_{jct}-\bar a^{-}_{jct}\big),\\
\bar a^{\pm}_{jct}&=\frac{1}{|\mathcal A^{\pm}_{jc}|}\sum_{x\in\mathcal A^{\pm}_{jc}}a_t(x),
\end{aligned}\tag{S2.1}\label{eq:S2.1}
\end{equation*}
where \ensuremath{\mathcal{C}}(\textit{t}) are the admissible classes of \textit{t} and $w_c$ weights each class by its atlas size. The sign of \textit{d}ⱼₜ attaches \textit{t} to a pole. The null distribution of \textit{d}ⱼₜ comes from 2000 permutations of atlas membership within each class, and the false-discovery rate (FDR) is controlled over all axes and admissible terms per seed \cite{Benjamini1995}. A term names the pole of its sign if it is significant (FDR \textless{} 0.05) and if the same contrast on validation images has the same sign with \textit{p} \textless{} 0.05. Poles are named by their strongest kept terms, and a pole without one is described by its verified boundary cue or held-out measurement below (Tables \hyperref[tab:S2]{S2} and \hyperref[tab:S4]{S4}).

Conditioning on the class is what keeps the names from restating the class profiles, since the most typical images of a pole come from one or two classes, and their contrast names the diagnosis. Naming a dictionary element with a vision–language model follows \cite{CLIPDissect,DNCBM}. However, the named object is an exact component of the decision from our pipeline, and the VLM only labels it.

\paragraph{Axis descriptors, boundary cues, and measurements.} A kept term is an \textit{axis descriptor} if it holds both within and between classes. Within classes, it points to its pole in at least two classes and never to the other pole. Between classes, it separates the two classes that its pole favors most from the two that the other pole favors most, with an AUROC of at least 0.7. A \textit{boundary cue} reads a term between classes. For each pair of a − pole class X and a + pole class Y whose medians differ by at least 0.5 s.d. along the axis, it compares the 20\% of X furthest toward + with the 20\% of Y furthest toward −. Cues are read only for pairs that include one of the term's expected classes, class names are excluded, and a cue found on both poles is dropped. \textit{Held-out measurements} are image statistics and recorded variables, contrasted on validation images between 20\% within-class pole atlases in the same way. They include lesion size, color, and texture on skin, nuclear and cytoplasm features on PBC and BMC, view position, film brightness, and framing on chest, and the sub-class and acquisition variables (maturation stage, merged child classes, histopathology confirmation, crowding).

\begin{figure*}[p]
\centering
\phantomsection\label{fig:S1}
\includegraphics[width=\textwidth,height=0.86\textheight,keepaspectratio]{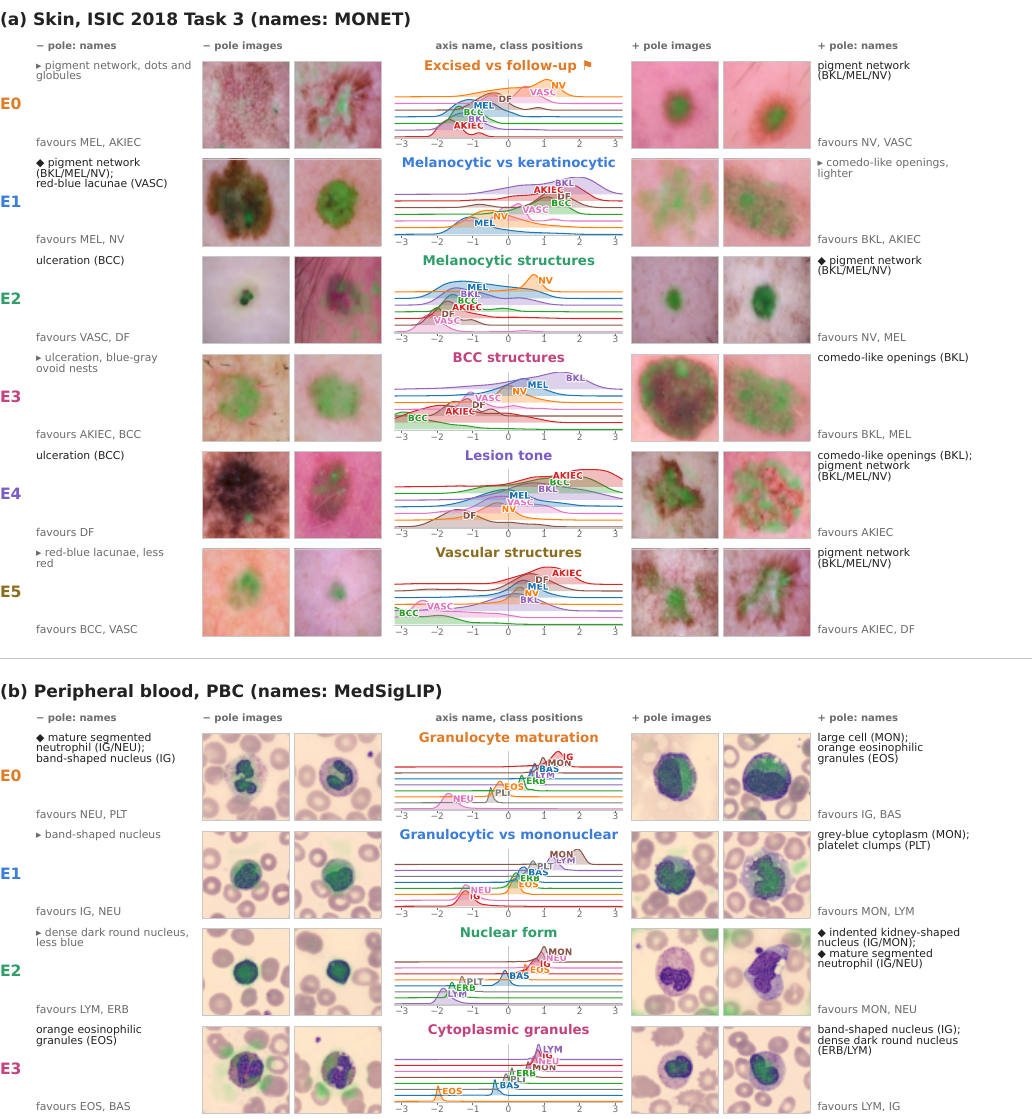}
\caption*{\textbf{Figure S1 (part 1 of 2).} Caption on the next page.}
\end{figure*}

\paragraph{Axis names.} Naming covers the four public datasets, whose modalities medical vision–language models are trained on, and the OBM axes enter every other analysis unnamed. The theme of an axis comes from the first tier that holds. T1 is a sub-class, acquisition, or context variable that is the axis's strongest held-out effect (at least 0.5 s.d.), or an axis that orders classes forming one stage series (Spearman \textbar{}ρ\textbar{} ≥ 0.6, as for blast to segmented neutrophil on BMC). T2 is a lineage split, where the two poles' differential classes fall into different lineage groups (for example, melanocytic against keratinocytic on skin). T3 is the theme of the axis descriptors, T4 that of the boundary cues, T3m a measurement descriptor that holds both within and between classes, and T5 the strongest held-out measurement. A name holds one theme, and each pole is described by its lineage group or classes and its two strongest names, or else by its strongest verified cue (Table \hyperref[tab:S2]{S2}, Fig. \hyperref[fig:S1]{S1}).

\paragraph{Acquisition flag.} Five axes whose strongest kept held-out measurements are acquisition or context variables (\textbar{}d\textbar{} ≥ 0.5) are flagged: skin E0 (more excised lesions toward −, 1.58 s.d.), BMC E2 (more crowded toward +, 0.90), chest E0 and E2 (more AP films toward −, 1.64 and 1.66) and chest E4 (more black film corners toward +, 1.30). A flag is not a shortcut verdict, and whether the classifier relies on a flagged variable is tested in Sec. \hyperref[sec:S9.7]{S9.7}. Because view position cannot vary within PA films, we also run the frozen protocol on PA films alone. There E2 is named hyperinflation (boundary cue), and E4 (film rotation) is flagged (Table \hyperref[tab:S4]{S4}, last rows, and main Fig. 1 and Sec. 5.4).

\paragraph{Frozen protocol.} One script with every rule and threshold fixed in code runs all datasets, and each permutation test draws from its own random stream, so a rerun reproduces every output exactly. Every seed-dependent quantity (names, differential, class medians, boundary pairs) is a consensus over the three seeds.

\begin{table*}[t]
\centering
\phantomsection\label{tab:S2}\caption*{\textbf{Table S2. Differential and pole names of each axis.} Consensus of three seeds. Kept VLM terms come first (axis descriptors in bold), then boundary cues (▸), and a pole with neither lists its held-out measurements. The last six rows read the chest basis on PA films only (main Fig. 1).}
\scriptsize
\setlength{\tabcolsep}{3pt}
\begin{tabularx}{\textwidth}{l>{\raggedright\arraybackslash}X>{\raggedright\arraybackslash}X>{\raggedright\arraybackslash}X}
\toprule
Axis & Differential (− ↔ +) & − pole names & + pole names \\
\midrule
Skin E0 & MEL, AKIEC ↔ NV, VASC & ▸ pigment network; ▸ dots and globules & pigment network; ▸ red-blue lacunae \\
Skin E1 & MEL, NV ↔ BKL, AKIEC & \textbf{pigment network}; red-blue lacunae & ▸ comedo-like openings \\
Skin E2 & VASC, DF ↔ NV, MEL & ulceration; ▸ red-blue lacunae & \textbf{pigment network}; ▸ pigment network; ▸ dots and globules \\
Skin E3 & AKIEC, BCC ↔ BKL, MEL & ▸ ulceration; ▸ blue-gray ovoid nests & comedo-like openings \\
Skin E4 & DF ↔ AKIEC & ulceration & comedo-like openings; pigment network \\
Skin E5 & BCC, VASC ↔ AKIEC, DF & ▸ red-blue lacunae & pigment network \\
PBC E0 & NEU, PLT ↔ IG, BAS & band-shaped nucleus; indented kidney-shaped nucleus; \textbf{mature segmented neutrophil}; ▸ platelet clumps; ▸ band-shaped nucleus & large cell; orange eosinophilic granules; ▸ large cell \\
PBC E1 & IG, NEU ↔ MON, LYM & ▸ band-shaped nucleus & gray-blue cytoplasm; platelet clumps; ▸ gray-blue cytoplasm; ▸ dense dark round nucleus \\
PBC E2 & LYM, ERB ↔ MON, NEU & ▸ dense dark round nucleus & \textbf{indented kidney-shaped nucleus}; gray-blue cytoplasm; \textbf{mature segmented neutrophil}; ▸ band-shaped nucleus \\
PBC E3 & EOS, BAS ↔ LYM, IG & orange eosinophilic granules; ▸ orange eosinophilic granules & band-shaped nucleus; dense dark round nucleus; indented kidney-shaped nucleus; ▸ dense dark round nucleus \\
BMC E0 & PEB, PMO ↔ ART, NIF & coarse purple granules; ▸ coarse purple granules & measured: smaller cell, darker chromatin \\
BMC E1 & PLM, MON ↔ EOS, NGS & measured: rounder nucleus & ▸ orange eosinophilic granules \\
BMC E2 & ART ↔ NIF & measured: smaller cell, yellower cytoplasm & orange eosinophilic granules \\
BMC E3 & NIF, PMO ↔ LYT, BAS & coarse purple granules & orange eosinophilic granules; ▸ dark basophilic granules \\
BMC E4 & NGB, NGS ↔ BLA, PLM & coarse purple granules; orange eosinophilic granules & measured: higher N:C, larger nucleus \\
BMC E5 & MON, MYB ↔ PEB, PLM & dark basophilic granules; \textbf{coarse purple granules} & no verified cue \\
Chest E0 & Effusion, Atelectasis ↔ Infiltration, Pneumonia & airspace consolidation; visible pleural line without lung markings; lung opacity; enlarged cardiac silhouette; blunting of the costophrenic angle; basal opacity with a meniscus sign; hyperinflated lungs with flattened diaphragms; perihilar haziness & clear lungs; ▸ lung opacity; ▸ basal opacity with a meniscus sign; ▸ blunting of the costophrenic angle \\
Chest E1 & Infiltration, Pneumonia ↔ Cardiomegaly, Mass & perihilar haziness; clear lungs & visible pleural line without lung markings; lung opacity \\
Chest E2 & Infiltration, Cardiomegaly ↔ Emphysema, Pneumothorax & airspace consolidation; enlarged cardiac silhouette & visible pleural line without lung markings; ▸ visible pleural line without lung markings \\
Chest E3 & Emphysema, Effusion ↔ Cardiomegaly, Hernia & visible pleural line without lung markings; blunting of the costophrenic angle; ▸ basal opacity with a meniscus sign; ▸ lung opacity; ▸ blunting of the costophrenic angle; ▸ enlarged cardiac silhouette; ▸ hyperinflated lungs with flattened diaphragms & perihilar haziness \\
Chest E4 & Emphysema ↔ Consolidation & clear lungs; ▸ hyperinflated lungs with flattened diaphragms & lung opacity; basal opacity with a meniscus sign; blunting of the costophrenic angle \\
Chest E5 & Hernia, Cardiomegaly ↔ Nodule, Mass & clear lungs; ▸ enlarged cardiac silhouette & airspace consolidation; lung opacity; perihilar haziness; basal opacity with a meniscus sign; hyperinflated lungs with flattened diaphragms; enlarged cardiac silhouette \\
Chest (PA films) E0 & Effusion, Atelectasis ↔ Infiltration, Pneumonia & visible pleural line without lung markings; \textbf{airspace consolidation}; basal opacity with a meniscus sign; blunting of the costophrenic angle; perihilar haziness; enlarged cardiac silhouette; hyperinflated lungs with flattened diaphragms & clear lungs; ▸ basal opacity with a meniscus sign; ▸ airspace consolidation; ▸ blunting of the costophrenic angle \\
Chest (PA films) E1 & Infiltration, Pneumonia ↔ Cardiomegaly, Mass & clear lungs & visible pleural line without lung markings; blunting of the costophrenic angle; basal opacity with a meniscus sign \\
Chest (PA films) E2 & Infiltration, Cardiomegaly ↔ Emphysema, Hernia & clear lungs; ▸ enlarged cardiac silhouette; ▸ hyperinflated lungs with flattened diaphragms & visible pleural line without lung markings; blunting of the costophrenic angle; hyperinflated lungs with flattened diaphragms; basal opacity with a meniscus sign; airspace consolidation; enlarged cardiac silhouette \\
Chest (PA films) E3 & Emphysema, Effusion ↔ Cardiomegaly, Hernia & visible pleural line without lung markings; blunting of the costophrenic angle; ▸ basal opacity with a meniscus sign; ▸ blunting of the costophrenic angle; ▸ enlarged cardiac silhouette & enlarged cardiac silhouette \\
Chest (PA films) E4 & Emphysema ↔ Consolidation & clear lungs; ▸ hyperinflated lungs with flattened diaphragms & basal opacity with a meniscus sign; blunting of the costophrenic angle \\
Chest (PA films) E5 & Hernia, Cardiomegaly ↔ Nodule, Mass & clear lungs; ▸ enlarged cardiac silhouette & \textbf{airspace consolidation}; basal opacity with a meniscus sign; hyperinflated lungs with flattened diaphragms; blunting of the costophrenic angle \\
\bottomrule
\end{tabularx}
\end{table*}

\begin{figure*}[p]
\centering
\includegraphics[width=\textwidth,height=0.88\textheight,keepaspectratio]{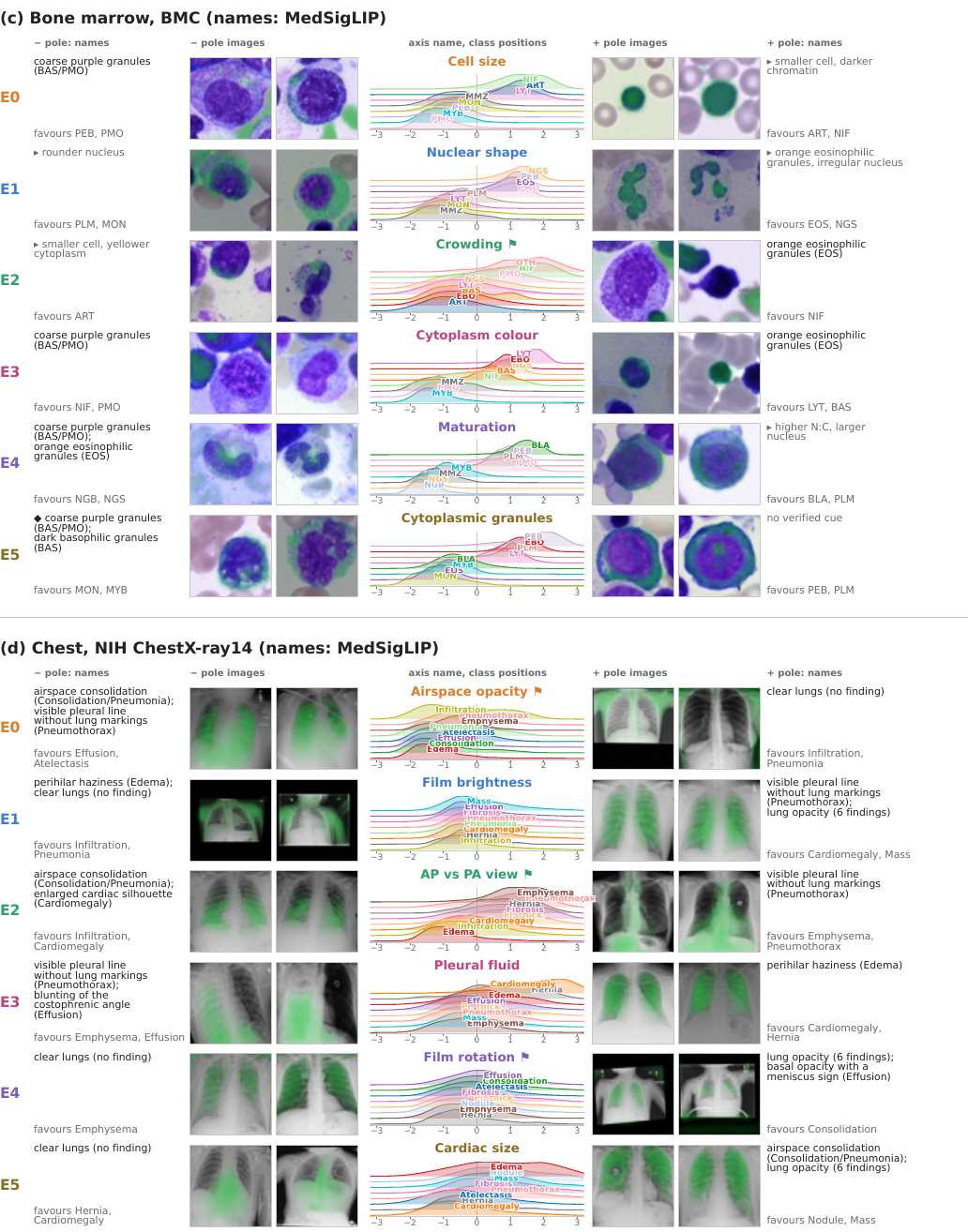}
\caption*{\textbf{Figure S1. Axis atlases of the four public datasets: (a) skin, (b) PBC, (c) BMC, and (d) chest.} Each row is one axis. Each pole shows its two most typical validation images with the pole map (green), its two strongest names (◆ axis descriptor, ▸ verified cue), and the classes its evidence favors (consensus of three seeds). The center gives the axis name (Table \hyperref[tab:S4]{S4}, with its ⚑ mark) above the validation densities of up to eight classes along the axis. Names come from MONET on skin and MedSigLIP elsewhere, and every image is correctly classified.}
\end{figure*}

\begin{table*}[t]
\centering
\phantomsection\label{tab:S3}\caption*{\textbf{Table S3. Names checked against annotations.} Consensus of three seeds. The reference is the ISIC 2018 Task 2 masks on skin and the maturation stages on PBC.}
\footnotesize
\setlength{\tabcolsep}{3pt}
\begin{tabularx}{\textwidth}{ll>{\raggedright\arraybackslash}X>{\raggedright\arraybackslash}Xl}
\toprule
Dataset, model & Axis, pole & Name & Reference & Verdict \\
\midrule
Skin, MONET & E0 + & pigment network & Task 2 masks: mask contrast −0.09 (toward −) & wrong sign \\
Skin, MONET & E1 − & pigment network & Task 2 masks: mask contrast −0.37 (toward −) & agrees \\
Skin, MONET & E2 + & pigment network & Task 2 masks: mask contrast +0.12 (toward +) & agrees \\
Skin, MONET & E4 + & pigment network & Task 2 masks: mask contrast +0.08 (toward +) & agrees \\
Skin, MONET & E5 + & pigment network & Task 2 masks: mask contrast +0.40 (toward +) & agrees \\
PBC, MedSigLIP & E0 − & indented kidney-shaped nucleus & stage order within NEU, IG & agrees \\
PBC, MedSigLIP & E0 + & large cell & stage order within NEU, IG & agrees \\
PBC, MedSigLIP & E0 − & mature segmented neutrophil & stage order within NEU, IG & agrees \\
PBC, MedSigLIP & E3 + & band-shaped nucleus & stage order within IG & agrees \\
PBC, MedSigLIP & E3 + & indented kidney-shaped nucleus & stage order within IG & agrees \\
\bottomrule
\end{tabularx}
\end{table*}

\begin{table*}[t]
\centering
\phantomsection\label{tab:S4}\caption*{\textbf{Table S4. Axis labels.} Consensus of three seeds, named by the precedence rule of Sec. \hyperref[sec:S2]{S2}. ⚑ marks an acquisition or context variable. Measurements are within-class pole contrasts on validation images (s.d.), boundary cues are contrasts between two classes, and ρ is the Spearman correlation of class medians. The last six rows use PA films only (main Fig. 1 and Sec. 5.4).}
\scriptsize
\setlength{\tabcolsep}{3pt}
\begin{tabularx}{\textwidth}{l>{\raggedright\arraybackslash}X>{\raggedright\arraybackslash}X}
\toprule
Axis & Name (tier) & Main support \\
\midrule
Skin E0 & \textbf{Excised vs follow-up} ⚑ (acquisition variable) & histopathology-confirmed share −1.58 s.d. toward +; boundary AKIEC vs VASC: red-blue lacunae +2.68 \\
Skin E1 & \textbf{Melanocytic vs keratinocytic} (lineage split) & descriptor pigment network (−; BKL/MEL/NV; AUROC 0.71); boundary NV vs BKL: comedo-like openings +0.53; lesion contrast −0.88 s.d. \\
Skin E2 & \textbf{Melanocytic structures} (axis descriptor) & descriptor pigment network (+; BKL/MEL/NV; AUROC 0.81); boundary VASC vs NV: red-blue lacunae −1.75; lesion lightness +0.70 s.d. \\
Skin E3 & \textbf{BCC structures} (boundary cue) & boundary BCC vs BKL: ulceration −1.01; BCC vs MEL: blue-gray ovoid nests −0.84 \\
Skin E4 & \textbf{Lesion tone} (measurement descriptor) & lesion lightness +0.24 s.d., ρ +0.54; lesion texture +0.50 s.d. \\
Skin E5 & \textbf{Vascular structures} (boundary cue) & boundary VASC vs AKIEC: red-blue lacunae −2.35; VASC vs DF −1.77; lesion size +0.93 s.d. \\
PBC E0 & \textbf{Granulocyte maturation} (sub-class order) & later stages toward − (IG −1.55, NEU −1.14 s.d.); descriptor mature segmented neutrophil (−; IG/NEU; AUROC 0.75) \\
PBC E1 & \textbf{Granulocytic vs mononuclear} (lineage split) & boundary NEU vs LYM: band-shaped nucleus −1.94; NEU vs MON: gray-blue cytoplasm +1.07 \\
PBC E2 & \textbf{Nuclear form} (axis descriptor) & descriptors indented kidney-shaped nucleus (+; IG/MON; AUROC 0.84) and mature segmented neutrophil (+; IG/NEU; 0.93) \\
PBC E3 & \textbf{Cytoplasmic granules} (boundary cue) & boundary EOS vs LYM: orange eosinophilic granules −1.07; EOS vs IG −0.92 \\
BMC E0 & \textbf{Cell size} (measurement descriptor) & cell radius −0.56 s.d., ρ −0.66; chromatin darkness +0.85 s.d. \\
BMC E1 & \textbf{Nuclear shape} (measurement descriptor) & nuclear circularity −0.39 s.d., ρ −0.61; boundary PLM vs EOS: orange eosinophilic granules +0.98 \\
BMC E2 & \textbf{Crowding} ⚑ (context variable) & crowding +0.90 s.d.; cytoplasm yellowness −0.87 s.d. \\
BMC E3 & \textbf{Cytoplasm color} (measurement descriptor) & cytoplasm yellowness +0.25 s.d., ρ +0.68; boundary NIF vs BAS: dark basophilic granules +1.55 \\
BMC E4 & \textbf{Maturation} (stage series) & stage order BLA → NGS ρ 0.80–0.82; N:C +0.54 s.d., ρ +0.79; nucleus radius +0.53 s.d., ρ +0.70 \\
BMC E5 & \textbf{Cytoplasmic granules} (axis descriptor) & descriptor coarse purple granules (−; BAS/PMO; AUROC 0.76) \\
Chest E0 & \textbf{Airspace opacity} ⚑ (boundary cue) & AP share −1.64 s.d. (flag); boundary Effusion vs Infiltration: lung opacity +1.80 \\
Chest E1 & \textbf{Film brightness} (measurement) & brightness +1.45, contrast −1.11, AP share +1.07 s.d. \\
Chest E2 & \textbf{AP vs PA view} ⚑ (acquisition variable) & AP share −1.66 s.d.; age +0.85 s.d. \\
Chest E3 & \textbf{Pleural fluid} (boundary cue) & boundary Effusion vs Cardiomegaly: meniscus sign −1.01, lung opacity −0.94 \\
Chest E4 & \textbf{Film rotation} ⚑ (acquisition variable) & black film corners +1.30 s.d.; brightness −1.00 s.d. \\
Chest E5 & \textbf{Cardiac size} (boundary cue) & boundary Cardiomegaly vs Nodule: enlarged cardiac silhouette −1.48 \\
Chest (PA films) E0 & \textbf{Airspace opacity} (axis descriptor) & descriptor airspace consolidation (−; Consolidation/Pneumonia; AUROC 0.72); boundary Effusion vs Pneumonia: basal opacity with a meniscus sign +1.64 \\
Chest (PA films) E1 & \textbf{Film brightness} (measurement) & brightness +1.91 s.d. toward +; contrast −0.79 s.d. toward + \\
Chest (PA films) E2 & \textbf{Hyperinflation} (boundary cue) & boundary Cardiomegaly vs Hernia: enlarged cardiac silhouette −1.28; boundary Cardiomegaly vs Emphysema: enlarged cardiac silhouette −0.91; age +1.47 s.d. toward + \\
Chest (PA films) E3 & \textbf{Pleural fluid} (boundary cue) & boundary Effusion vs Cardiomegaly: basal opacity with a meniscus sign −1.21; boundary Effusion vs Cardiomegaly: blunting of the costophrenic angle −1.18 \\
Chest (PA films) E4 & \textbf{Film rotation} ⚑ (acquisition variable) & black film corners +1.49 s.d. toward +; boundary Emphysema vs Consolidation: hyperinflated lungs with flattened diaphragms −0.90 \\
Chest (PA films) E5 & \textbf{Cardiac size} (boundary cue) & descriptor airspace consolidation (+; Consolidation/Pneumonia; AUROC 0.71); boundary Cardiomegaly vs Nodule: enlarged cardiac silhouette −1.11 \\
\bottomrule
\end{tabularx}
\end{table*}

\section{Content codes and descriptor residual}\label{sec:S3}

\paragraph{Content codes (main Sec. 3.4).} Each concept token is mapped into a shared space, $y_u=Q_{l(u)}[\mathrm{LN}(t_u);\bar c_u]+q_{l(u)}$, where $\bar c_u$ are standardized CIELAB statistics of the concept's region. They restore the color that the self-supervised encoder discards, since it is trained to be color-invariant \cite{LooC}. The content code of axis $j$ is the signed, presence-weighted average of its mapped concept tokens $y_u$,
\begin{equation*}
z_j(x)=P_j\,\frac{\sum_u(\pi^{+}_{uj}-\pi^{-}_{uj})\,g_u(x)\,y_u(x)}{\sum_u(\pi^{+}_{uj}+\pi^{-}_{uj})\,g_u(x)}. \tag{S3.1}\label{eq:S3.1}
\end{equation*}
A coherence term $\mathcal{L}_{\text{coh}}$ keeps the tokens of a pole close to their mean \cite{ProtoPNet}, and a redundancy term $\mathcal{L}_{\text{red}}$ penalizes the label-conditional $R^2$ of each code predicted linearly from the others \cite{Schmidhuber1992}. Content codes condition only the generator (Sec. \hyperref[sec:S4]{S4}) and never enter main Eq. (6).

\paragraph{Descriptor residual (main Sec. 3.4).} General context, such as the acquisition appearance that differs between devices and sites \cite{Bissoto2019,Bissoto2020}, gets its own descriptor residual so that it does not leak into the axis codes. The descriptor residual is read from all concepts and received by the generator as one token group. Part of it is anchored to the recorded image transformations (color cast, tone, and zoom) \cite{AugSelf}. On transformed training pairs, the change of each anchored part must predict its transformation parameter while the other parts stay put ($\mathcal{L}_{\text{anc}}$), and the rest is free to hold any remaining context. A hinge $\mathcal{L}_\eta$ caps its label-explained variance, so it cannot become a second, uninterpretable decision channel \cite{Havasi2022}, and it never enters main Eq. (6). A light broadcast decoder \cite{DINOSAUR} reconstructs the concept tokens from the content codes and the descriptor residual ($\mathcal{L}_{\text{rec}}$), which keeps both informative for the generator.

\paragraph{Objective.} The concept basis is learned by minimizing
\begin{equation*}
\begin{aligned}
\mathcal{L}&=\mathcal{L}_{\text{basis}}+\mathcal{L}_{\text{content}},\\
\mathcal{L}_{\text{basis}}&=\lambda_{\text{pole}}\mathcal{L}_{\text{pole}}+\lambda_{\text{vol}}\mathcal{L}_{\text{vol}}+\lambda_{\text{bal}}\mathcal{L}_{\text{bal}}\\
&\quad+\lambda_{\text{cmp}}\mathcal{L}_{\text{cmp}}+\lambda_{\text{agr}}\mathcal{L}_{\text{agr}},\\
\mathcal{L}_{\text{content}}&=\mathcal{L}_{\text{rec}}+\lambda_{\text{coh}}\mathcal{L}_{\text{coh}}+\lambda_{\text{red}}\mathcal{L}_{\text{red}}\\
&\quad+\lambda_{\text{anc}}\mathcal{L}_{\text{anc}}+\lambda_{\eta}\mathcal{L}_{\eta},
\end{aligned}\tag{S3.2}\label{eq:S3.2}
\end{equation*}
The basis term gives the axes the four properties of main Sec. 3.3 (Sec. \hyperref[sec:S1]{S1}), and the content term makes them inspectable and keeps acquisition appearance in the descriptor residual. Exactness and accuracy need no term, because they hold by construction (main Eq. (6)).

\section{Generator details}\label{sec:S4}

\paragraph{Condition tokens and edits.} The generator is conditioned only on the concept basis and the descriptor residual, and the identity of the image comes from inversion. Each axis $j$ gives one condition token from its content code $z_j(x)$ (Eq. \hyperref[eq:S3.1]{(S3.1)}) and its globally standardized position $\tilde p_j(x)$,
\begin{equation*}
c_j(x)=A\,z_j(x)+a\,\tilde p_j(x)+\eta_j, \tag{S4.1}\label{eq:S4.1}
\end{equation*}
where $A$ is a learned projection, $a$ a learned vector, and $\eta_j$ a learned embedding that identifies axis $j$. The $K$ axis tokens and the descriptor residual (Sec. \hyperref[sec:S3]{S3}) form the condition $c(x)$. A learned Gaussian layer smooths the condition space between classes, and every inversion, edit, and interpolation uses its mean. To edit an image along axis $j$ by $\delta$ within-class standard deviations $\sigma_j$, we change its condition to
\begin{equation*}
c'(x)=c(x)+\mathcal{P}\big(p(x)+\delta\sigma_j e_j\big)-\mathcal{P}\big(p(x)\big), \tag{S4.2}\label{eq:S4.2}
\end{equation*}
where $p(x)$ is the vector of axis positions, $e_j$ the $j$-th unit vector, and $\mathcal P(p)$ the kernel-weighted mean condition of training images of the same class near position $p$. The edit thus moves toward real images further along the axis. The class, the image's own label, only selects the neighbors.

\paragraph{Generator.} We train a rectified flow \cite{RectifiedFlow,FlowMatching} in pixel space. With noise $\epsilon\sim\mathcal N(0,I)$ and $x_t=(1-t)\,\epsilon+t\,x$, a U-Net \cite{UNet,ADM} $v_\psi(x_t,t,c)$ predicts the velocity $x-\epsilon$ and reads the condition $c(x)$ by cross-attention at every resolution \cite{LDM}. Training images of cells and skin lesions are flipped and rotated but keep their colors, because their orientation carries no diagnostic information whereas color does. Chest films are neither flipped nor rotated, as this would make anatomically impossible images.

\paragraph{Joint training of basis and generator.} The content codes are learned to reconstruct concept tokens, not pixels, so they may lack detail that a renderer needs. We therefore fine-tune the basis and the descriptor residual together with the generator at a ten times smaller learning rate, while the encoder and the concept bottleneck stay frozen. An anchor keeps the basis at the solution learned in main Sec. 3.3,
\begin{equation*}
\begin{aligned}
\mathcal L_{\text{gen}}={}&\mathbb E_{x,\epsilon,t}\big\|v_\psi\big(x_t,t,c(x)\big)-(x-\epsilon)\big\|^2\\
&+\lambda_{\text{a}}\,\mathbb E_x\Big(\big\|\tilde p(x)-\tilde p^{\,0}(x)\big\|^2+\frac{\|z(x)-z^{0}(x)\|^2}{\mathrm{Var}(z^{0})}\Big),
\end{aligned}\tag{S4.3}\label{eq:S4.3}
\end{equation*}
where $\tilde p^{\,0}$ and $z^0$ are the positions and codes before fine-tuning, and $z$ collects the content codes and the descriptor residual. Fine-tuning cannot cost accuracy or faithfulness, because the bottleneck is frozen and main Eq. (6) holds for every assignment, so the fine-tuned basis still decomposes the same logits exactly. The anchor protects the properties that the flow loss does not see: distinctness, balance, coherence, and reproducibility. No class label or logit enters the generator or its loss. The U-Net has base width 32 and is trained with batch 32, learning rate $10^{-4}$, and an EMA of 0.9995 for 30k steps, with anchor weight $\lambda_a=1$. Learned null tokens replace single condition tokens with probability 0.15 and whole token groups with 0.3, and the Gaussian layer is added in a short fine-tune with KL weight $10^{-3}$.

\paragraph{Rendering by inversion.} To keep the identity of the image, we invert it fully to noise under its own condition $c(x)$ with Heun's method \cite{EDM}, as in deterministic diffusion inversion \cite{DDIM}. We then regenerate it from this noise under the edited condition $c'$ with delta guidance of weight $w$ toward $c'$ and away from $c(x)$,
\begin{equation*}
\begin{aligned}
\hat\epsilon&=\Phi_{1\to0}\big[v_\psi(\cdot,\cdot,c(x))\big](x),\\
\tilde v&=v_\psi(\cdot,\cdot,c')\\
&\quad+(w-1)\big(v_\psi(\cdot,\cdot,c')-v_\psi(\cdot,\cdot,c(x))\big),\\
x'&=\Phi_{0\to1}[\tilde v](\hat\epsilon),
\end{aligned}\tag{S4.4}\label{eq:S4.4}
\end{equation*}
where $\Phi_{a\to b}[v]$ integrates the ODE $\dot x_t=v(x_t,t)$ from time $a$ to $b$. With $c'=c(x)$ the guidance term vanishes and the procedure returns the reconstruction $\hat x$. Every edit is compared with $\hat x$ rather than with $x$, so that inversion error is not counted as an effect of the edit. The kernel of $\mathcal P$ widens from bandwidth 0.5 until its effective sample size reaches 30, and edits use 60 function evaluations of Heun's method, delta guidance $w=4$, and steps of $\pm2$ within-class s.d.

\paragraph{Checking an edit.} An edit is faithful if the image moves along the requested axis by the requested amount and along no other. Two readers measure the change from $\hat x$ to $x'$. The frozen model and its basis re-classify the edit and give the shift of every position and of the margin, and an independent reader that shares no weights with the encoder gives its own reading of the positions, which we compare with its change between real cells one within-class s.d. apart. A random direction of the same size in position space, rendered in the same way, serves as control (Sec. \hyperref[sec:S9]{S9}, Table \hyperref[tab:S11]{S11}). The generator and the edits do not enter main Eq. (6). The explanation of a prediction remains the exact decomposition, and a rendered image only illustrates one of its axes.

\section{Datasets and accuracy}\label{sec:S5}

\subsection{Datasets}\label{sec:S5.1}

\paragraph{Datasets.} All five datasets are class-imbalanced, and the public ones keep their official test splits where they exist.

\paragraph{Skin.} ISIC 2018 Task 3 \cite{HAM10000,ISIC2018} consists of dermoscopic images of pigmented skin lesions from the HAM10000 collection, acquired at the Medical University of Vienna and at a skin cancer practice in Queensland, Australia. More than half of the diagnoses are confirmed by histopathology and the rest by follow-up, expert consensus or confocal microscopy. The seven classes are melanoma (MEL), melanocytic nevus (NV), basal cell carcinoma (BCC), actinic keratosis and intraepithelial carcinoma (AKIEC), benign keratosis (BKL, covering solar lentigo, seborrheic keratosis and lichen planus-like keratosis), dermatofibroma (DF), and vascular lesion (VASC). The largest class (NV) outnumbers the smallest (DF) by 58 to 1. We test on the official 1,512 test images. The official validation set holds only 193 images, so we pool it with the 10,015 training images and hold out 15\% of each class for validation, grouped by lesion, as one lesion often has several photographs. The 193 official validation images carry no lesion identifier and stay in training, which leaves 8,674 training and 1,534 validation images.

\paragraph{PBC.} The peripheral blood cell dataset of \citet{Acevedo2020} holds 17,092 images of single normal blood cells in Giemsa-stained peripheral blood smears. They were captured with a CellaVision DM96 analyzer at the Hospital Clinic of Barcelona from healthy donors. The eight classes are neutrophil (NEU), eosinophil (EOS), basophil (BAS), lymphocyte (LYM), monocyte (MON), immature granulocyte (IG), erythroblast (ERB) and platelet (PLT). The largest class (NEU) outnumbers the smallest (LYM) by 2.7 to 1. With no official split or patient identifiers, we split it at random within each class into 11,959 training, 1,705 validation and 3,428 test images (70/10/20).

\paragraph{OBM.} OBM is a private, IRB-approved grayscale dataset of 40,920 stain-free normal white blood cells in five classes, basophil (BAS), eosinophil (EOS), lymphocyte (LYM), monocyte (MON) and neutrophil (NEU). The largest class outnumbers the smallest by 1.6 to 1. Oblique back-illumination microscopy images the unstained cells with phase contrast from ex vivo microfluidic experiments. The dataset is split by imaging session that gives 28,527 training, 4,183 validation and 8,210 test images (70/10/20).

\paragraph{BMC.} The bone marrow cytology dataset of \citet{Matek2021} holds 171,374 expert-annotated single-cell images from Giemsa/Pappenheim-stained bone marrow smears of 945 patients with a variety of hematological diseases, imaged by 40× bright-field microscopy. The 21 classes cover three cell lineages, immature or abnormal cells, and non-diagnostic categories. The myeloid lineage holds the neutrophil series, promyelocyte (PMO), myelocyte (MYB), metamyelocyte (MMZ), band neutrophil (NGB) and segmented neutrophil (NGS), together with eosinophil (EOS), abnormal eosinophil (ABE), basophil (BAS) and monocyte (MON). The erythroid lineage holds proerythroblast (PEB) and erythroblast (EBO). The lymphoid lineage holds lymphocyte (LYT), immature lymphocyte (LYI), plasma cell (PLM) and hairy cell (HAC). Blast (BLA) and faggot cell with Auer rod bundles (FGC) are immature or abnormal cells, and smudge cell (KSC), other cell (OTH), artifact (ART) and not identifiable (NIF) are non-diagnostic. The largest class (NGS, 29,424 images) outnumbers the smallest (ABE, 8 images) by about 3,700 to 1. The release has no official split. One image (MYB\_05527) is truncated in the release and excluded, and we split the remaining images at random within each class into 70\% training, 10\% validation and 20\% test images (119,950, 17,129 and 34,294). The release has no patient identifiers, so, as on PBC, the split is per image.

\paragraph{Chest.} NIH ChestX-ray14 \cite{ChestXray14} holds 112,120 frontal chest radiographs (67,310 posteroanterior, PA, and 44,810 anteroposterior, AP) of 30,805 patients from the NIH Clinical Center. Its 14 thoracic findings were mined from the radiology reports by natural language processing, so the labels carry some noise. The findings are atelectasis, cardiomegaly, consolidation, edema, effusion, emphysema, fibrosis, hernia, infiltration, mass, nodule, pleural thickening, pneumonia and pneumothorax. The most frequent finding (infiltration, 19,894 images) occurs 88 times as often as the rarest (hernia, 227 images). An image can carry several findings or none, and an image without findings (53.8\% of all images, 58.4\% of training and 38.5\% of test images) is negative for all 14. We keep the official patient-wise test set (25,596 images of 2,797 patients) and split the official training-and-validation list (86,524 images of 28,008 patients) by patient into 76,488 training and 10,036 validation images (24,881 and 3,127 patients), with the same distribution of images per patient and the same prevalence of each finding in both splits.

\paragraph{Evaluation-only annotations.} Five kinds of annotation serve only as evaluation ground truth and never enter training. The ISIC 2018 Task 2 attribute masks \cite{ISIC2018} outline five dermoscopic structures on 2,594 images disjoint from HAM10000, with one binary mask per structure and image. Pigment network is present in 58.7\% of the images, milia-like cysts in 26.3\%, globules in 23.2\%, negative network in 7.3\% and streaks in 3.9\%. We use the 2,590 lesions diagnosed as nevus, melanoma or seborrheic keratosis. The HAM10000 lesion segmentations \cite{HAM10000} give one binary lesion mask for each of the 10,015 training images. The PBC maturation-stage labels are recorded in the file names, promyelocyte (PMY), myelocyte (MY) and metamyelocyte (MMY) among the immature granulocytes, and band (BNE) and segmented (SNE) among the neutrophils. The NIH bounding boxes \cite{ChestXray14} mark eight chest findings with 984 boxes on 880 test images. The fifth is the acquisition metadata of each dataset.

\subsection{Accuracy}\label{sec:S5.2}

\paragraph{Baseline protocols.} We retrain every code-released method that needs no concept labels, select checkpoints and hyperparameters on validation, and score each in its final interpretable form. The baselines fall into four families: \textbf{1. The frozen readouts} are the linear probe on the last layer \cite{DINOv2}, the 12-layer vote of per-layer linear probes, the attentive probe \cite{AIM} and Head2Toe \cite{Head2Toe}, all on the same frozen DINOv3-B as our model. The cross-layer fusion baselines of main Table 2b, ScalarMix \cite{ELMo}, VFusion \cite{VFusion}, ALF \cite{ALF} and single linear heads on the concatenated CLS tokens or the concatenated CLS‖mean features of all layers, read the frozen DINOv3-B like the probes; \textbf{2. Among the prototype and compact-head models}, PNP \cite{PNP} and PIP-Net \cite{PIPNet} read the frozen DINOv3-B, as each falls below the linear probe on at most one of the five datasets (main Table 1). MCPNet \cite{MCPNet} and ProtoViT \cite{ProtoViT} fell far below the linear probe on a frozen encoder, so they are retrained end to end with their original protocols, MCPNet on its ResNet-50 and ProtoViT on DINOv3-B; \textbf{3. The text-defined concept models}, LF-CBM \cite{LFCBM}, PCBM-h \cite{PCBM} and Explicd \cite{Explicd}, need a vision–language encoder and read frozen MedSigLIP \cite{MedGemma}. \textbf{4. Among the learned-concept models}, DN-CBM \cite{DNCBM} also reads frozen MedSigLIP, and BotCL \cite{BotCL} is trained end to end on DINOv3-B for the same reason as MCPNet and ProtoViT.

\begin{table*}[t]
\centering
\phantomsection\label{tab:S5a}\caption*{\textbf{Table S5a. Training settings of the Table 1 baselines.} Checkpoints are chosen on validation. Hyperparameters are tuned on validation where the table says so, and otherwise follow the official settings.}
\scriptsize
\setlength{\tabcolsep}{3pt}
\begin{tabularx}{\textwidth}{>{\raggedright\arraybackslash}X>{\raggedright\arraybackslash}Xl>{\raggedright\arraybackslash}X>{\raggedright\arraybackslash}X>{\raggedright\arraybackslash}X>{\raggedright\arraybackslash}X}
\toprule
Family & Method & Encoder & Trained parts & Schedule & Learning rate & Code \\
\midrule
Frozen readout & Linear probe (last layer), attentive probe & DINOv3-B, frozen & head only & 30 epochs & grid, chosen on validation & ours, a standard head trained side by side with the others under the same protocol as our concept bottleneck \\
 & Linear probe, 12-layer vote & DINOv3-B, frozen & one linear head per layer & 30 epochs & grid, chosen on validation & ours, one linear head per layer, no reference code needed \\
 & Head2Toe & DINOv3-B, frozen & group-lasso feature selection + linear head & 10 epochs selection + 30 epochs head & 1e-4; sparsity, fraction and weight decay chosen on validation & PyTorch port, as the official code is JAX for other backbones \\
Cross-layer fusion (main Table 2b) & ScalarMix, VFusion, ALF, concatenated CLS and CLS‖mean heads & DINOv3-B, frozen & head only & 30 epochs & grid, chosen on validation & ours, from the papers (no released VFusion code, ScalarMix exists only for NLP), under the same protocol as our concept bottleneck \\
Prototype / compact head & PNP & DINOv3-B, frozen & part prototypes + head & official, stage 2 ≈ 540k images & official & port of the official code \\
 & PIP-Net & DINOv3-B, frozen & prototypes + head & official medical schedule, ≈ 2k + 10k updates & official & port of the official code \\
 & MCPNet & ResNet-50, fine-tuned & all & 50 epochs & 1e-4 & official code \\
 & ProtoViT & DINOv3-B, fine-tuned & all & official & official & official code \\
Text-defined concepts & LF-CBM, PCBM-h & MedSigLIP, frozen & concept layer + sparse linear head & closed form & grid, chosen on validation & port of the official code \\
 & Explicd & MedSigLIP, frozen & criteria prompts + head & fixed update budget (150 epochs on skin) & official & official code, chest criteria map ours (Sec. \hyperref[sec:S5.2]{S5.2}) \\
Learned concepts & DN-CBM & MedSigLIP, frozen & concept layer + sparse linear head & closed form & grid, chosen on validation & port of the official code \\
 & BotCL & DINOv3-B, fine-tuned & all & official & official & official code \\
\bottomrule
\end{tabularx}
\end{table*}

\paragraph{Multi-label adaptations for chest (‡ in Table 1).} PNP, MCPNet and Explicd assume one class per image, whereas a chest radiograph carries any subset of the 14 findings, or none. We change only their label handling and keep the rest of their official recipes. All three predict 15 outputs (the 14 findings and No Finding), train them with binary cross-entropy and are scored by the macro AUROC over the 14 findings. In PNP \cite{PNP}, each foreground patch is assigned to the positive finding of its image whose prototypes it matches best, and each finding receives a free bias that starts at its prevalence. In MCPNet \cite{MCPNet}, each finding has a positive and a negative mean concept distribution, its score is the distance to the negative one minus the distance to the positive one, and the class-concept loss weights positives by 1/(2p) and negatives by 1/(2(1−p)) for a finding of prevalence p. For Explicd \cite{Explicd}, we wrote a chest criteria map in the format of the official skin map, and an image's target for a criterion is the uniform distribution over the values that its positive findings name, or the normal value when none does.

\paragraph{Cross-layer fusion baselines (main Sec. 5.1 and Table 2b).} Head2Toe keeps 10 to 20\% of the 55,296 pooled features of all 12 layers. Averaged over the five datasets, Head2Toe, ScalarMix, VFusion and ALF fall 7.5 to 12.6 points below our model (Table \hyperref[tab:S6]{S6}). Main Table 2b condenses the six cross-layer readouts of Table \hyperref[tab:S6]{S6} into one row. On each dataset it reports the readout with the best validation score, which is Head2Toe on skin and chest, the head on the concatenated CLS tokens on PBC and OBM, and VFusion on BMC, and the test set is read only for that readout, as for every other model in the paper.

\paragraph{Domain-specific SOTA (main Table 1, row v).} Each range spans the best published results on the same public dataset. We take the paper's score in our metric, balanced accuracy or the macro AUROC over the 14 chest findings, recompute it from published per-class results when the paper reports another metric, and otherwise use the reported accuracy. On skin, both results use the official test set. \citet{Shen2022} reach 85.3 with the cleanest protocol, and \citet{Yao2022} reach 87.5 with test-time augmentation and model selection on the test set. PBC has no official split. MedMNIST+ \cite{MedMNISTPlus} reports 99.0 accuracy on its fixed split of the same images, and 98.9 is the balanced accuracy recomputed from the per-class test counts of \citet{Firat2024}. On BMC, the dataset paper \cite{Matek2021} and the DinoBloom linear probe \cite{DinoBloom} both reach 71.4. For Matek et al., this is the balanced accuracy over the 20 classes (except abnormal eosinophils) recomputed from its per-class recalls, and DinoBloom uses its own 80/20 split. The LoRA-adapted DINOv3 of CytoDINO \cite{CytoDINO} reaches 75.7, also on its own 80/20 split. On chest, all three results use the official patient-wise split to fine tune the encoder, after masked pretraining on unlabeled radiographs \cite[82.3]{Xiao2023}, with an attention ensemble at 224 px \cite[82.6]{A3Net}, or at 768 px after pretraining on about 700,000 radiographs \cite[84.4]{ArkPlus}. OBM is private and has no published reference. We excluded results evaluated on training images or on a validation split, and results on substantially different label sets. These differences make the PBC and BMC ranges indicative.

\section{Ablations of pipeline designs on all five datasets}\label{sec:S6}

\subsection{Within a layer (main Table 2a)}\label{sec:S6.1}

\paragraph{Measures.} \textbf{Mean accuracy} is the test balanced accuracy of one layer's head (macro AUROC on chest), averaged over the 12 layers and 3 seeds. The other measures validation and are averaged over layers. The core of a concept's attention map is the smallest set of its 196 patches that holds half the map's mass. \textbf{Attention coverage} is the share of patches in at least one of the 12 cores, and \textbf{attention overlap} is the mean pairwise IoU of the cores. \textbf{Slot similarity} is the mean pairwise within-class linear centered kernel alignment (CKA) \cite{Kornblith2019} of the 12 slot tokens. \textbf{Shared errors} are the share of test errors that two layers at least four apart make on the same images, averaged over all such pairs. \textbf{Seed agreement} measures whether the concepts across seeds attend to the same places. The 12 concepts between seeds are paired one to one by Hungarian match on the correlation of their attention maps. Seed agreement is the mean correlation of the paired maps minus that of a random pairing, so it is near zero when the concepts are interchangeable, for example when all of them attend to the same region.

\begin{table}[tbp]
\centering
\phantomsection\label{tab:S5}\caption*{\textbf{Table S5. Main Table 2a on all five datasets.} Best per dataset in bold, second underlined.}
\small
\setlength{\tabcolsep}{4pt}
\resizebox{\columnwidth}{!}{%
\begin{tabular}{@{}lccccc@{}}
\toprule
\textbf{Arm} & \makecell{\textbf{Mean}\\\textbf{acc.}\,$\uparrow$} & \makecell{\textbf{Att. coverage}\,$\uparrow$\\\textbf{(overlap}\,$\downarrow$\textbf{)}} & \makecell{\textbf{Slot}\\\textbf{similarity}\,$\downarrow$} & \makecell{\textbf{Shared}\\\textbf{errors}\,$\downarrow$} & \makecell{\textbf{Seed}\\\textbf{agreement}\,$\uparrow$} \\
\midrule
\multicolumn{6}{@{}l}{\textbf{\emph{Skin}}} \\
Probe heads (CLS‖mean) & 67.1 & \textbf{1.00} (—) & — & 0.43 & — \\
MQA (n = 12) & 73.8 & 0.26 (0.58) & 0.50 & \uline{0.39} & 0.01 \\
Slot attention (SA, n = 12) & 74.5 & \textbf{1.00} (0.22) & \textbf{0.45} & \uline{0.39} & 0.48 \\
SA + covariance & \textbf{76.4} & \uline{0.99} (\uline{0.17}) & 0.59 & 0.40 & \textbf{0.66} \\
\textbf{+ gates (ours)} & \uline{75.0} & \uline{0.99} (\textbf{0.16}) & \uline{0.48} & \textbf{0.37} & \uline{0.60} \\
\addlinespace[2pt]
\multicolumn{6}{@{}l}{\textbf{\emph{PBC}}} \\
Probe heads (CLS‖mean) & 95.9 & \textbf{1.00} (—) & — & \textbf{0.23} & — \\
MQA (n = 12) & 98.6 & 0.12 (0.62) & \uline{0.28} & 0.27 & 0.03 \\
Slot attention (SA, n = 12) & 98.6 & \textbf{1.00} (0.32) & \textbf{0.15} & 0.27 & 0.29 \\
SA + covariance & \textbf{98.8} & \uline{0.99} (\uline{0.19}) & 0.30 & 0.30 & \textbf{0.53} \\
\textbf{+ gates (ours)} & \uline{98.7} & 0.98 (\textbf{0.14}) & 0.29 & \uline{0.26} & \uline{0.51} \\
\addlinespace[2pt]
\multicolumn{6}{@{}l}{\textbf{\emph{OBM}}} \\
Probe heads (CLS‖mean) & 62.7 & \textbf{1.00} (—) & — & \uline{0.50} & — \\
MQA (n = 12) & 67.4 & 0.41 (0.67) & 0.51 & 0.59 & 0.02 \\
Slot attention (SA, n = 12) & \textbf{70.0} & \textbf{1.00} (0.34) & \textbf{0.40} & 0.59 & 0.40 \\
SA + covariance & 69.7 & \uline{0.99} (\uline{0.30}) & 0.48 & 0.59 & \uline{0.55} \\
\textbf{+ gates (ours)} & \uline{69.8} & \uline{0.99} (\textbf{0.25}) & \uline{0.42} & \textbf{0.43} & \textbf{0.56} \\
\addlinespace[2pt]
\multicolumn{6}{@{}l}{\textbf{\emph{BMC}}} \\
Probe heads (CLS‖mean) & 58.7 & \textbf{1.00} (—) & — & \uline{0.44} & — \\
MQA (n = 12) & 72.1 & 0.29 (0.56) & 0.23 & \uline{0.44} & 0.06 \\
Slot attention (SA, n = 12) & 73.1 & \uline{0.99} (0.19) & \textbf{0.19} & \uline{0.44} & \uline{0.52} \\
SA + covariance & \uline{73.6} & 0.98 (\textbf{0.07}) & 0.29 & 0.46 & \textbf{0.61} \\
\textbf{+ gates (ours)} & \textbf{73.7} & 0.98 (\uline{0.08}) & \uline{0.22} & \textbf{0.43} & \textbf{0.61} \\
\addlinespace[2pt]
\multicolumn{6}{@{}l}{\textbf{\emph{Chest}}} \\
Probe heads (CLS‖mean) & 73.3 & \textbf{1.00} (—) & — & n/a & — \\
MQA (n = 12) & 76.5 & 0.09 (0.77) & 0.53 & n/a & 0.00 \\
Slot attention (SA, n = 12) & 76.7 & \textbf{1.00} (0.32) & \textbf{0.42} & n/a & 0.49 \\
SA + covariance & \uline{76.8} & \textbf{1.00} (\uline{0.23}) & 0.58 & n/a & \uline{0.65} \\
\textbf{+ gates (ours)} & \textbf{78.3} & \uline{0.99} (\textbf{0.20}) & \uline{0.50} & n/a & \textbf{0.70} \\
\bottomrule
\end{tabular}}
\end{table}

\subsection{Across depth (main Table 2b)}\label{sec:S6.2}

\begin{figure*}[t]
\centering
\phantomsection\label{fig:S2}
\includegraphics[width=\textwidth,height=0.5\textheight,keepaspectratio]{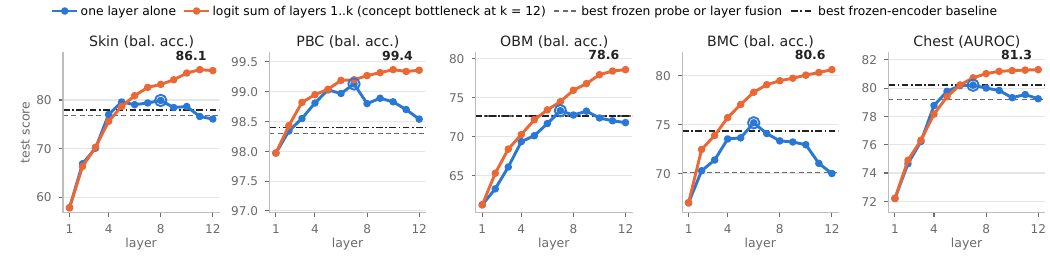}
\caption*{\textbf{Figure S2. The best single layer lies mid-depth, and summing the evidence of all layers climbs above every one of them.} Test score of each layer's head alone (blue, with a ring on the peak layer) and of the logit sum of layers 1 to k (orange), against the best frozen probe or layer fusion of main Table 1 (dashed) and the best frozen-encoder baseline of any family in main Table 1 (dash-dot).}
\end{figure*}

\paragraph{Methods.} Feature fusion, one head, merges the layers before classification. Six readouts fuse the frozen encoder features of all 12 layers into a single head (Sec. \hyperref[sec:S5.2]{S5.2}). The other rows merge after classification. On the 12 per-layer heads of every final concept bottleneck, we score each layer alone, the cumulative logit sum from shallow to deep, and five merge rules: the layer with the best validation score, the probability average, non-negative layer weights fitted on validation data by class-balanced cross-entropy, forward greedy layer selection on validation, and a class-balanced logistic regression on the concatenated layer logits fitted on validation. At least one layer classifies the image correctly for 95.9\% of skin, 99.8\% of PBC, and 94.8\% of BMC test images. Learned layer weights put most of their mass on intermediate layers (on OBM, 0.13 to 0.19 each on layers 5 to 9 and at most 0.02 on layers 1 to 4), but never improve on the equal-weight sum. Greedy layer selection is the one rule that scores above the sum, by 0.2 points on chest only. It picks its layers on validation, falls 0.2 to 1.5 points below the sum on the other four datasets, and keeps only a subset of the layers, so it gives up the all-layer decomposition of main Eq. (3). Table \hyperref[tab:S6]{S6} lists every rule, and Fig. \hyperref[fig:S2]{S2} every layer.

\begin{table*}[t]
\centering
\phantomsection\label{tab:S6}\caption*{\textbf{Table S6. Every readout across depth (full main Table 2b).} Test, mean ± s.d. over 3 seeds, with the best score per dataset in bold and the second best underlined. The feature-fusion rows fuse the encoder features of all layers, give the validation score in brackets, and italicize the readout that main Table 2b reports on each dataset, chosen on validation.}
\footnotesize
\setlength{\tabcolsep}{3pt}
\begin{tabularx}{\textwidth}{>{\raggedright\arraybackslash}Xlllll}
\toprule
Readout across depth & Skin & PBC & OBM & BMC & Chest \\
\midrule
Best single layer & 80.0 ± 0.2 & 99.0 ± 0.1 & 73.3 ± 0.3 & 74.7 ± 0.4 & 80.2 ± 0.2 \\
Feature fusion: Head2Toe (main Table 1) & \textit{75.2 ± 0.5 (79.6)} & 97.6 ± 0.3 (98.4) & 72.7 ± 0.3 (63.2) & 65.5 ± 0.7 (69.1) & \textit{77.4 ± 0.4 (79.7)} \\
Feature fusion: ScalarMix & 73.7 ± 0.8 (78.6) & 97.5 ± 0.2 (97.7) & 69.8 ± 0.1 (61.7) & 64.3 ± 0.2 (64.5) & 75.4 ± 0.0 (77.2) \\
Feature fusion: VFusion & 70.8 ± 0.4 (72.0) & 98.4 ± 0.1 (98.8) & 70.6 ± 0.4 (64.7) & \textit{65.1 ± 0.4 (71.1)} & 68.4 ± 0.3 (71.6) \\
Feature fusion: ALF & 72.0 ± 0.5 (76.4) & 96.5 ± 0.1 (96.4) & 63.1 ± 0.1 (58.8) & 58.4 ± 0.3 (59.9) & 73.0 ± 0.1 (75.5) \\
Feature fusion: single head on the concatenated CLS & 69.2 ± 1.0 (71.9) & \textit{98.3 ± 0.1 (99.2)} & \textit{71.3 ± 0.3 (64.8)} & 68.7 ± 0.4 (66.4) & 73.9 ± 0.0 (75.3) \\
Feature fusion: single head on the concatenated CLS‖mean & 68.7 ± 0.3 (71.8) & 98.2 ± 0.1 (98.9) & 69.8 ± 0.3 (62.8) & 64.1 ± 0.2 (65.9) & 70.1 ± 0.1 (74.8) \\
Probability average & 84.3 ± 0.3 & \uline{99.3 ± 0.0} & 76.0 ± 0.2 & \uline{79.9 ± 1.4} & 81.2 ± 0.0 \\
Stacked classifier on layer logits & 75.0 ± 0.7 & 98.9 ± 0.1 & 74.2 ± 1.3 & 66.0 ± 0.1 & 75.7 ± 0.1 \\
Greedy layer subset & 84.6 ± 0.4 & 99.2 ± 0.1 & 77.4 ± 0.5 & 79.4 ± 1.3 & \textbf{81.5 ± 0.0} \\
Learned layer weights (on the layer logits) & \uline{85.6 ± 0.5} & 99.2 ± 0.1 & \uline{77.7 ± 0.6} & 79.4 ± 1.2 & 80.8 ± 0.0 \\
\textbf{Logit sum, Eq. (3)} & \textbf{86.1 ± 0.2} & \textbf{99.4 ± 0.0} & \textbf{78.6 ± 0.3} & \textbf{80.6 ± 0.3} & \uline{81.3 ± 0.0} \\
\bottomrule
\end{tabularx}
\end{table*}

\subsection{Concept basis (main Table 2c)}\label{sec:S6.3}

\paragraph{Measures.} Table 2c reports four properties of the final basis on validation images, averaged over the three bases, and Table \hyperref[tab:S7]{S7} gives them for every dataset together with the decision statistics discussed below. \textbf{Distinctness} is the \uline{number of effective axes}, the participation ratio (Σλ)²/Σλ² of the eigenvalues of the within-class correlation matrix of the \textit{K} positions. \textbf{Compactness} is the \uline{number of axes per decision}, the fewest axes whose contributions reach 80\% of an image's positive margin without the class bias, with the number of concepts needed for the same share in brackets. \textbf{Reproducibility} is the \uline{seed match}. The axes of two bases are paired one to one by Hungarian matching on the absolute correlation of their positions, and we report the worst-matched axis with the median in brackets. \textbf{Stability} is the \uline{within-class correlation of each position between clean and augmented views} of the same training images, with the same measure for the concepts in brackets.

\paragraph{Configuration selection.} Before seeing any result of the final recipe, we fixed six validation bars. The basis must be exact (F1). Its positions must have an effective dimension of at least 0.6\textit{K} and a pairwise \textbar{}corr\textbar{} of at most 0.6 (F2). No axis may hold more than half of the average margin, and every axis must hold more than 5\% of it (F3). The median seed match must reach 0.9 (F4). Content redundancy, the within-class \textit{R}² of each axis's content code predicted from the others, must stay at or below 0.5 (F5). Readability, the balanced accuracy of a logistic regression on the \textit{K} positions, may fall at most one point below that of the unconstrained basis (F6). Configurations are ranked by bars passed, then by the smallest unique variance of any axis (1 − \textit{R}² from the other \textit{K} − 1), then by readability, and then by smaller \textit{K}.

\begin{table*}[t]
\centering
\phantomsection\label{tab:S7}\caption*{\textbf{Table S7. The basis on all five datasets (full main Table 2c).}}
\scriptsize
\setlength{\tabcolsep}{3pt}
\begin{tabularx}{\textwidth}{ll>{\raggedright\arraybackslash}Xll>{\raggedright\arraybackslash}X>{\raggedright\arraybackslash}X>{\raggedright\arraybackslash}X>{\raggedright\arraybackslash}X>{\raggedright\arraybackslash}X}
\toprule
Dataset & Effective axes ↑ & Axes per decision ↓ (concepts) & Seed match ↑ (median) & Stability ↑ (concepts) & Decisions on 1 / 2 / 3 / ≥ 4 axes (\%) & Most used axis, margin share (\%) & Axes leading ≥ 5\% of decisions & Distinct class leaders (classes) & Least needed axis, in 80\% set (\%) \\
\midrule
Skin & 5.42/6 & 2.0 (26.0) & 0.89 (0.96) & 0.82 (0.72) & 6.9 / 69.5 / 22.8 / 0.8 & 30.6 & 5 & 5 (5) & 11.6 \\
PBC & 3.95/4 & 1.7 (21.3) & 0.99 (0.99) & 0.96 (0.86) & 40.7 / 23.0 / 34.6 / 1.6 & 33.4 & 4 & 4 (8) & 23.1 \\
OBM & 3.88/4 & 1.0 (20.3) & 0.98 (0.98) & 0.78 (0.60) & 78.6 / 19.8 / 1.6 / 0.0 & 35.0 & 4 & 4 (5) & 6.3 \\
BMC & 5.62/6 & 2.7 (27.3) & 0.85 (0.90) & 0.97 (0.92) & 7.5 / 40.3 / 41.7 / 10.5 & 32.4 & 3 & 3.7 (17) & 16.5 \\
Chest & 5.61/6 & 1.0 (15.3) & 0.88 (0.95) & 0.93 (0.81) & 75.3 / 15.3 / 9.0 / 0.4 & 32.8 & 4 & 5 (10) & 6.2 \\
\bottomrule
\end{tabularx}
\end{table*}

\paragraph{Axes for each decision (main Sec. 5.2).} We record the number of axes every decision rests on in validation and test sets. The leading axis set is the smallest set of axes that carries 80\% of its positive margin (Table \hyperref[tab:S7]{S7}). At least 89.5\% of decisions rest on three axes or fewer on every dataset, whereas the bottleneck needs a median of 15 to 27 concepts for the same share. One or two axes suffice for 98\% of OBM and 91\% of chest decisions, while skin and BMC decisions mostly rest on two or three axes (92\% and 82\%).

\paragraph{No axis dominates the margin.} The most used axis carries 31 to 35\% of the margin on average, and three to five axes each lead at least 5\% of the decisions. Different classes are led by different axes. On skin, melanoma is led by E1 (melanocytic vs keratinocytic) in 93\% of its decisions, nevus by E2 in 86\%, basal cell carcinoma by E5 in 100\%, and benign keratosis by E3 in 96\%. On PBC, E1 (granulocytic vs mononuclear) leads immature granulocytes and monocytes, and E0 (granulocyte maturation) leads neutrophils.

\paragraph{Together the decisions need every axis.} An axis that rarely leads still contributes to many decisions. On every dataset, each axis lies in the 80\% set of at least 6\% of decisions.

\paragraph{Stable leaders and audit suspects.} Leading axes are stable across seeds on skin, OBM, and chest, where every class keeps its leader in all three seeds. On PBC, five of eight classes keep it, and the other three switch between two axes. On BMC, the maturation axis E4 of main Fig. 4 leads 54\% of all decisions, because several BMC classes are stages of one maturation series. The leaders also point the acquisition audit of Sec. \hyperref[sec:S9.7]{S9.7} to its suspects. AKIEC is led by the flagged axis E0 (excised vs follow-up) in 68\% of its decisions, yet the audit finds that the classifier does not rely on E0 (AUROC gap 0.020 with and against the prior). On chest, eight findings are led by E0, E1, or E2, which track view position and film brightness besides opacity. The audit confirms the known view-position confound on E0 and E2, the one reliance it finds and the one main Sec. 5.4 reports, and clears film rotation on E4.

\section{Generalization to other foundation-model backbones}\label{sec:S7}

\paragraph{Backbones.} Each backbone replaces the frozen DINOv3-B/16 in the full pipeline (the concept bottleneck, then its gated fine-tune), with all other settings unchanged, and is scored for accuracy only. DINOv2 ViT-B/14 reads 196 px images, so its 14 × 14 patch grid matches DINOv3-B/16 at 224 px, and all 12 layers are read. SAM2 Hiera-B+, the image encoder of the SAM2 segmentation model, has 24 blocks in four stages and reads 224 px images. Every second block is read at its own grid and width, from 56 × 56 in the first stage to 7 × 7 in the last. MedSigLIP, a medical vision–language encoder with 27 layers, reads 448 px images. Twelve evenly spaced layers from the fifth onward are read, with their 32 × 32 patch grids area-pooled to 14 × 14. Its pretraining includes chest radiographs (MIMIC-CXR) and dermatology images, but none of our five datasets is among its listed training sets \cite{MedGemma}.

\paragraph{Results (Table \hyperref[tab:S8]{S8}).} The gains of our pipeline generalize to other foundation-model backbones. With every backbone on every dataset, the 12-layer logit sum beats the best single layer by up to 6.1 points, except for a tie for SAM2 on PBC (99.0), and it lies 1.8 to 29.9 points above the last-layer linear probe. The gain is largest where the last layer is weakest, so the spread between the four backbones shrinks from 9.0 to 27.3 points under the linear probe to 0.4 to 8.2 points under the logit sum. MedSigLIP gives the best last-layer probe on every dataset, as it is pretrained on medical images, yet DINOv3 gives the best logit sum on skin, OBM, and BMC and ties MedSigLIP on PBC (99.4). MedSigLIP leads on chest (82.5 vs 81.3), where it reads 448 px images, twice the resolution of DINOv3, and the logit sum adds 1.9 points over its best layer. With frozen encoders, the logit sum thus lies within the domain-specific SOTA range on skin (DINOv3) and chest (MedSigLIP) matches it on PBC, and lies above it on BMC.

\begin{table}[tbp]
\centering
\phantomsection\label{tab:S8}\caption*{\textbf{Table S8. The logit sum with other frozen backbones.} Linear probe reads the last layer, Best layer is the single layer chosen on validation, and Logit sum is main Eq. (3). Test, mean ± s.d. over 3 seeds. The last row repeats the domain-specific SOTA ranges of main Table 1, each under its own protocol (Sec. \hyperref[sec:S5.2]{S5.2}).}
\footnotesize
\setlength{\tabcolsep}{4pt}
\sbox\fittab{\begin{tabular}{@{}lccccc@{}}
\toprule
Readout & \makecell{Skin} & \makecell{PBC} & \makecell{OBM} & \makecell{BMC} & \makecell{Chest} \\
\midrule
\multicolumn{6}{@{}l}{\textit{DINOv3-B/16 (main paper)}} \\
Linear probe & 71.1 ± 0.2 & 96.6 ± 0.1 & 62.5 ± 0.3 & 59.4 ± 0.2 & 75.2 ± 0.0 \\
Best layer & 80.0 ± 0.2 & 99.0 ± 0.1 & 73.3 ± 0.3 & 74.7 ± 0.4 & 80.2 ± 0.2 \\
Logit sum & \textbf{86.1 ± 0.2} & \textbf{99.4 ± 0.0} & \textbf{78.6 ± 0.3} & \textbf{80.6 ± 0.3} & \uline{81.3 ± 0.0} \\
\addlinespace[2pt]
\multicolumn{6}{@{}l}{\textit{DINOv2-B/14}} \\
Linear probe & 69.8 ± 0.4 & 96.8 ± 0.2 & 58.4 ± 1.4 & 62.1 ± 0.4 & 75.2 ± 0.2 \\
Best layer & 79.9 ± 0.3 & 98.8 ± 0.0 & 68.2 ± 1.2 & 77.6 ± 0.2 & 79.7 ± 0.6 \\
Logit sum & \uline{84.5 ± 0.3} & \uline{99.3 ± 0.2} & 72.6 ± 0.6 & 80.0 ± 0.4 & 80.9 ± 0.4 \\
\addlinespace[2pt]
\multicolumn{6}{@{}l}{\textit{SAM2 Hiera-B+}} \\
Linear probe & 50.2 ± 0.6 & 88.6 ± 0.4 & 56.7 ± 0.3 & 48.4 ± 0.7 & 69.9 ± 0.2 \\
Best layer & 75.9 ± 0.4 & 99.0 ± 0.1 & 66.8 ± 0.8 & 73.8 ± 0.6 & 78.3 ± 0.4 \\
Logit sum & 80.1 ± 0.4 & 99.0 ± 0.1 & 70.4 ± 0.4 & 76.4 ± 0.2 & 78.7 ± 0.2 \\
\addlinespace[2pt]
\multicolumn{6}{@{}l}{\textit{MedSigLIP}} \\
Linear probe & 77.5 ± 0.2 & 97.6 ± 0.2 & 69.6 ± 0.5 & 68.1 ± 0.4 & 80.5 ± 0.2 \\
Best layer & 78.5 ± 0.3 & \uline{99.3 ± 0.1} & 72.4 ± 0.4 & 76.2 ± 0.4 & 80.6 ± 0.1 \\
Logit sum & 83.4 ± 0.2 & \textbf{99.4 ± 0.1} & \uline{77.4 ± 0.5} & \uline{80.3 ± 0.2} & \textbf{82.5 ± 0.2} \\
\addlinespace[2pt]
\multicolumn{6}{@{}l}{\textit{Domain-specific SOTA}} \\
Published & 85.3–87.5 & 98.9–99.0 & — & 71.4–75.7 & 82.3–84.4 \\
\bottomrule
\end{tabular}}%
\ifdim\wd\fittab>\columnwidth\resizebox{\columnwidth}{!}{\usebox\fittab}\else
\begin{tabular*}{\columnwidth}{@{\extracolsep{\fill}}lccccc}
\toprule
Readout & \makecell{Skin} & \makecell{PBC} & \makecell{OBM} & \makecell{BMC} & \makecell{Chest} \\
\midrule
\multicolumn{6}{@{}l}{\textit{DINOv3-B/16 (main paper)}} \\
Linear probe & 71.1 ± 0.2 & 96.6 ± 0.1 & 62.5 ± 0.3 & 59.4 ± 0.2 & 75.2 ± 0.0 \\
Best layer & 80.0 ± 0.2 & 99.0 ± 0.1 & 73.3 ± 0.3 & 74.7 ± 0.4 & 80.2 ± 0.2 \\
Logit sum & \textbf{86.1 ± 0.2} & \textbf{99.4 ± 0.0} & \textbf{78.6 ± 0.3} & \textbf{80.6 ± 0.3} & \uline{81.3 ± 0.0} \\
\addlinespace[2pt]
\multicolumn{6}{@{}l}{\textit{DINOv2-B/14}} \\
Linear probe & 69.8 ± 0.4 & 96.8 ± 0.2 & 58.4 ± 1.4 & 62.1 ± 0.4 & 75.2 ± 0.2 \\
Best layer & 79.9 ± 0.3 & 98.8 ± 0.0 & 68.2 ± 1.2 & 77.6 ± 0.2 & 79.7 ± 0.6 \\
Logit sum & \uline{84.5 ± 0.3} & \uline{99.3 ± 0.2} & 72.6 ± 0.6 & 80.0 ± 0.4 & 80.9 ± 0.4 \\
\addlinespace[2pt]
\multicolumn{6}{@{}l}{\textit{SAM2 Hiera-B+}} \\
Linear probe & 50.2 ± 0.6 & 88.6 ± 0.4 & 56.7 ± 0.3 & 48.4 ± 0.7 & 69.9 ± 0.2 \\
Best layer & 75.9 ± 0.4 & 99.0 ± 0.1 & 66.8 ± 0.8 & 73.8 ± 0.6 & 78.3 ± 0.4 \\
Logit sum & 80.1 ± 0.4 & 99.0 ± 0.1 & 70.4 ± 0.4 & 76.4 ± 0.2 & 78.7 ± 0.2 \\
\addlinespace[2pt]
\multicolumn{6}{@{}l}{\textit{MedSigLIP}} \\
Linear probe & 77.5 ± 0.2 & 97.6 ± 0.2 & 69.6 ± 0.5 & 68.1 ± 0.4 & 80.5 ± 0.2 \\
Best layer & 78.5 ± 0.3 & \uline{99.3 ± 0.1} & 72.4 ± 0.4 & 76.2 ± 0.4 & 80.6 ± 0.1 \\
Logit sum & 83.4 ± 0.2 & \textbf{99.4 ± 0.1} & \uline{77.4 ± 0.5} & \uline{80.3 ± 0.2} & \textbf{82.5 ± 0.2} \\
\addlinespace[2pt]
\multicolumn{6}{@{}l}{\textit{Domain-specific SOTA}} \\
Published & 85.3–87.5 & 98.9–99.0 & — & 71.4–75.7 & 82.3–84.4 \\
\bottomrule
\end{tabular*}\fi
\end{table}

\section{Structural evaluation of the concept basis}\label{sec:S8}

\paragraph{Basis properties (main Fig. 3).} \textbf{Exactness} is one minus the share of the logit margin that the axes leave unexplained. \textbf{Balance} is set by the largest and smallest average share of the margin held by an axis. \textbf{Completeness} is how much of the concept presences the axes predict, relative to PCA with as many components. \textbf{Distinctness} is the effective dimension of the within-class axis positions, rescaled to (effective axes − 1)/(\textit{K} − 1). \textbf{Coherence} is the pole score of Eq. \hyperref[eq:S1.3]{(S1.3)}. \textbf{Reproducibility} is the correlation of the worst-matched axis across jointly trained bases after Hungarian matching of their positions. All six lie in [0, 1] and are averaged over 3 bases.

\paragraph{Alternatives (Table \hyperref[tab:S9]{S9}).} We compare with 19 ways of extracting \textit{K} = 6 axes from the same concept presences and concept-to-class weights, drawn from the five families of main Fig. 3 and re-implemented on the same concept pool. \textbf{Factor rotation} partitions the concepts by PCA, varimax \cite{Rohe2023}, promax, sparse PCA, or ICA. \textbf{Correlation clustering} groups them by \textit{k}-means, Ward agglomeration, or a ProtoPShare-style merge \cite{ProtoPShare}. \textbf{Class-aware selection} keeps non-redundant concepts by mRMR \cite{mRMR}, a QPM-style quadratic selection \cite{QPM}, or an IBD-style decomposition \cite{IBD}. \textbf{Matrix factorization} and \textbf{sparse dictionaries} learn atoms with inferred codes, namely NMF as in CRAFT \cite{CRAFT}, semi-NMF, archetypal analysis \cite{ArchetypalSAE}, whitened PCA as in Concept Whitening \cite{ConceptWhitening}, matching-pursuit codes \cite{MPSAE}, and ReLU, TopK, and Matryoshka sparse autoencoders \cite{MatryoshkaSAE}. Partitions are exact by construction, while the other methods leave part of the margin in a residual.

\begin{table}[tbp]
\centering
\phantomsection\label{tab:S9}\caption*{\textbf{Table S9. All 19 condensations of main Fig. 3.} † Shown in main Fig. 3, the method of each family with the highest lowest score. Our basis is set in bold.}
\scriptsize
\setlength{\tabcolsep}{2pt}
\begin{tabularx}{\columnwidth}{>{\raggedright\arraybackslash}Xlcccccc}
\toprule
Family & Method & \rotatebox{90}{\shortstack[l]{Exactness ↑}} & \rotatebox{90}{\shortstack[l]{Balance ↑}} & \rotatebox{90}{\shortstack[l]{Completeness ↑}} & \rotatebox{90}{\shortstack[l]{Distinctness ↑}} & \rotatebox{90}{\shortstack[l]{Coherence ↑}} & \rotatebox{90}{\shortstack[l]{Reproducibility ↑}} \\
\midrule
Factor rotation & Varimax & 1.00 & 0.54 & 0.98 & 0.67 & 0.93 & 0.47 \\
 & Promax & 1.00 & 0.56 & 0.98 & 0.61 & 0.96 & 0.29 \\
 & ICA & 1.00 & 0.55 & 0.98 & 0.54 & 0.99 & 0.47 \\
 & Sparse PCA & 1.00 & 0.59 & 0.99 & 0.55 & 0.97 & 0.48 \\
 & PCA † & 1.00 & 0.53 & 0.98 & 0.65 & 0.94 & 0.68 \\
Correlation clustering & \textit{k}-means (ACE) & 1.00 & 0.59 & 0.97 & 0.43 & 0.98 & 0.73 \\
 & Ward & 1.00 & 0.54 & 0.95 & 0.39 & 0.99 & 0.69 \\
 & ProtoPShare merge † & 1.00 & 0.46 & 0.96 & 0.63 & 1.00 & 0.50 \\
Class-aware selection & QPM-lite & 1.00 & 0.56 & 0.95 & 0.46 & 0.99 & 0.50 \\
 & IBD-lite & 0.95 & 0.61 & 0.99 & 0.38 & 0.73 & 0.91 \\
 & mRMR † & 1.00 & 0.59 & 0.94 & 0.48 & 0.99 & 0.48 \\
Matrix factorization & NMF (CRAFT) & 0.95 & 0.57 & 0.98 & 0.66 & 0.72 & 0.53 \\
 & Semi-NMF & 0.98 & 0.73 & 0.99 & 0.53 & 0.60 & 0.55 \\
 & Whitened PCA & 0.57 & 0.56 & 1.00 & 0.86 & 0.46 & 0.77 \\
 & Archetypal analysis † & 0.89 & 0.80 & 0.91 & 0.74 & 0.59 & 0.65 \\
Sparse dictionaries & OMP (MP-SAE) & 0.89 & 0.55 & 0.89 & 0.71 & 0.72 & 0.54 \\
 & SAE TopK & 0.50 & 0.66 & 0.91 & 0.78 & 0.74 & 0.88 \\
 & SAE Matryoshka & 0.59 & 0.64 & 0.95 & 0.73 & 0.63 & 0.43 \\
 & SAE ReLU † & 0.66 & 0.64 & 0.98 & 0.64 & 0.63 & 0.87 \\
\textbf{Concept basis} & \textbf{Ours} & \textbf{1.00} & \textbf{0.80} & \textbf{0.98} & \textbf{0.91} & \textbf{0.78} & \textbf{0.83} \\
\bottomrule
\end{tabularx}
\end{table}

\paragraph{Ablations (Table \hyperref[tab:S10]{S10}).} Removing a term costs mainly the property it targets, scored as in main Fig. 3. Without the balance band, balance falls from 0.80 to 0.46. Without the volume term, the axes collapse onto a shared factor and distinctness falls from 0.91 to 0.39. Without coherent poles the axes stay distinct but stop being contrasts, as coherence falls from 0.78 to 0.29. Without the consensus the worst-matched axis across bases drops from 0.83 to 0.19, while the median axis still matches at 0.84.

\begin{table}[tbp]
\centering
\phantomsection\label{tab:S10}\caption*{\textbf{Table S10. Ablation of the basis terms (main Sec. 5.2).} Skin, scored as in main Fig. 3. Removing a term mainly breaks the property it protects, on the diagonal. Reproducibility is the worst-matched axis, with the median in brackets.}
\footnotesize
\setlength{\tabcolsep}{4pt}
\sbox\fittab{\begin{tabular}{@{}lcccc@{}}
\toprule
Variant & Balance ↑ & Distinctness ↑ & Coherence ↑ & Reproducibility ↑ (median) \\
\midrule
\textbf{Full recipe (main Fig. 3)} & 0.80 & 0.91 & 0.78 & 0.83 (0.95) \\
Without balance band & 0.46 & 0.87 & 0.92 & 0.60 (0.96) \\
Without volume & 0.79 & 0.39 & 0.99 & 0.93 (0.98) \\
Without coherent poles & 0.80 & 0.97 & 0.29 & 0.86 (0.94) \\
Without consensus & 0.77 & 0.92 & 0.78 & 0.19 (0.84) \\
\bottomrule
\end{tabular}}%
\ifdim\wd\fittab>\columnwidth\resizebox{\columnwidth}{!}{\usebox\fittab}\else
\begin{tabular*}{\columnwidth}{@{\extracolsep{\fill}}lcccc}
\toprule
Variant & Balance ↑ & Distinctness ↑ & Coherence ↑ & Reproducibility ↑ (median) \\
\midrule
\textbf{Full recipe (main Fig. 3)} & 0.80 & 0.91 & 0.78 & 0.83 (0.95) \\
Without balance band & 0.46 & 0.87 & 0.92 & 0.60 (0.96) \\
Without volume & 0.79 & 0.39 & 0.99 & 0.93 (0.98) \\
Without coherent poles & 0.80 & 0.97 & 0.29 & 0.86 (0.94) \\
Without consensus & 0.77 & 0.92 & 0.78 & 0.19 (0.84) \\
\bottomrule
\end{tabular*}\fi
\end{table}

\section{Extended results}\label{sec:S9}

\subsection{Position–margin agreement}\label{sec:S9.1}

\paragraph{Position tracks margin (main Sec. 5.2).} An axis's pole classes are the two groups of classes whose evidence its poles favor most. \textbf{Position–margin agreement} is the within-class Spearman ρ between an image's position on an axis and that axis's margin between its pole classes, taken as the median over axes and the mean over seeds. It is 0.93 on skin, 0.91 on PBC, 0.99 on OBM, 0.63 on BMC, and 0.79 on chest, 0.85 on average. Decorrelated axes can still depend on each other, so a position and its margin could both follow a factor that the other axes carry. Pairing each image with its nearest neighbor on the other axes reduces this concern (agreement within pairs 0.69 to 0.99).

\subsection{Maturation axes and counterfactual edits}\label{sec:S9.2}

\paragraph{Maturation order and edit check (main Sec. 5.3).} Each BMC stage is a separate, unordered training class, yet the medians of the granulocytic classes on E4 follow their maturation order (Kendall τ 0.87, permutation p 0.017 in every seed). We edit 96 PBC test images and 68 BMC validation images by ±2 within-class s.d. along each axis with Eq. \hyperref[eq:S4.2]{(S4.2)} and compare each edit with the reconstruction of its image (Table \hyperref[tab:S11]{S11}). Four measures check an edit. \textbf{Recon. LPIPS} is the perceptual distance between an image and its unedited reconstruction. With 0.0003 on PBC and 0.0011 on BMC, the generator reproduces each cell almost exactly, so what changes in an edit comes from the edit. \textbf{Selectivity} re-encodes every edit with the frozen concept bottleneck and basis and divides the shift of the edited axis, per requested s.d., by the mean absolute shift of the other axes. The edited axis moves 4.1 to 7.5 times more than the others on PBC and 7.4 to 14.3 times more on BMC, so an edit changes the requested axis and leaves the rest of the explanation in place. Because selectivity is measured by our own model, an independent reader checks the maturation edits. The reader is a ridge regression from ImageNet ConvNeXt-T features \cite{ConvNeXt} to the annotated maturation stage (promyelocyte to segmented neutrophil), fitted to the annotated stage rank on real training cells. It shares neither weights nor a training target with our model. On real validation cells its output follows that rank (Spearman ρ 0.91 on PBC and 0.86 on BMC). \textbf{Reader: maturation} is the change in the reader's output per requested s.d. of an edit. Between real cells one s.d. apart, the same reader changes by 0.33 on PBC and 0.30 on BMC. PBC edits move it by 0.36 (95\% CI 0.30 to 0.43), and BMC edits move it by 0.26 (0.18 to 0.35), close to what real maturation does. \textbf{Random control} applies an edit of the same size in a random direction of position space. It moves the reader by at most 0.07 on PBC and 0.04 on BMC, and leaves the stage reader unchanged (95\% CI includes zero), so both readers respond to the direction of an edit, not to any change of the image. Together, the maturation axes order cells without ordered stage labels, and edits along them move the requested axis and mature the cell for an independent reader, so the counterfactuals of main Fig. 4 render what each maturation axis encodes.

\begin{table}[tbp]
\centering
\phantomsection\label{tab:S11}\caption*{\textbf{Table S11. Counterfactual edits move the requested axis and keep the cell.} \textbf{Reader: maturation} is the ConvNeXt reader's change per requested s.d. (95\% CI), \textbf{Random control} is the same reader under an edit of the same size in a random direction, and \textbf{Selectivity} is the edited axis's shift per requested s.d. over the mean absolute shift of the other axes (range over axes).}
\footnotesize
\setlength{\tabcolsep}{4pt}
\sbox\fittab{\begin{tabular}{@{}lclcc@{}}
\toprule
\makecell{Dataset, split\\(images)} & \makecell{Recon.\\LPIPS} & \makecell{Reader:\\maturation} & \makecell{Random\\control} & \makecell{Selectivity} \\
\midrule
PBC, test (96) & 0.0003 & 0.36 (0.30–0.43) & ≤ 0.07 & 4.1–7.5 † \\
BMC, val. (68) & 0.0011 & 0.26 (0.18–0.35) & ≤ 0.04 & 7.4–14.3 ‡ \\
\bottomrule
\end{tabular}}%
\ifdim\wd\fittab>\columnwidth\resizebox{\columnwidth}{!}{\usebox\fittab}\else
\begin{tabular*}{\columnwidth}{@{\extracolsep{\fill}}lclcc}
\toprule
\makecell{Dataset, split\\(images)} & \makecell{Recon.\\LPIPS} & \makecell{Reader:\\maturation} & \makecell{Random\\control} & \makecell{Selectivity} \\
\midrule
PBC, test (96) & 0.0003 & 0.36 (0.30–0.43) & ≤ 0.07 & 4.1–7.5 † \\
BMC, val. (68) & 0.0011 & 0.26 (0.18–0.35) & ≤ 0.04 & 7.4–14.3 ‡ \\
\bottomrule
\end{tabular*}\fi
\par\vspace{2pt}{\footnotesize\raggedright † PBC axes E0 to E2 (granulocyte maturation, granulocytic vs mononuclear, nuclear form). ‡ BMC maturation axis E4.\par}
\end{table}

\subsection{Axis-based differentials for rare abnormal cells, with and without their labels}\label{sec:S9.3}

\paragraph{Protocol (main Sec. 5.3).} To hide the abnormal labels, we retrain the concept bottleneck and the basis with each abnormal class merged into the normal class it resembles (17 classes), hairy cells (HAC) and immature lymphocytes (LYI) into lymphocytes (LYT), faggot cells (FGC) into promyelocytes (PMO), and abnormal eosinophils (ABE) into eosinophils (EOS). Abnormal eosinophils (8 images, 3 in test) are too few to score. The merged model never sees the abnormal labels in training. They only choose one axis and its direction per abnormal class on the training images, and the untouched test images give its AUROC against the parent, with no classifier fitted (Table \hyperref[tab:S12]{S12}). The 21-class reference model is scored the same way.

\begin{table}[tbp]
\centering
\phantomsection\label{tab:S12}\caption*{\textbf{Table S12. Abnormal cells against their parent class on a training-selected axis} (AUROC, mean ± s.d. over three seeds). Bold marks the higher of the two AUROCs.}
\footnotesize
\setlength{\tabcolsep}{4pt}
\sbox\fittab{\begin{tabular}{@{}llcc@{}}
\toprule
Abnormal vs parent & n, rare vs. common & Separate labels & Merged parent labels \\
\midrule
Hairy cell vs lymphocyte & 409 vs. 26,242 & \textbf{0.86 ± 0.02} & 0.81 ± 0.01 \\
Immature vs mature lymphocyte & 65 vs. 26,242 & \textbf{0.98 ± 0.00} & 0.89 ± 0.02 \\
Faggot cell vs promyelocyte & 47 vs. 11,994 & 0.91 ± 0.03 & \textbf{0.92 ± 0.01} \\
\bottomrule
\end{tabular}}%
\ifdim\wd\fittab>\columnwidth\resizebox{\columnwidth}{!}{\usebox\fittab}\else
\begin{tabular*}{\columnwidth}{@{\extracolsep{\fill}}llcc}
\toprule
Abnormal vs parent & n, rare vs. common & Separate labels & Merged parent labels \\
\midrule
Hairy cell vs lymphocyte & 409 vs. 26,242 & \textbf{0.86 ± 0.02} & 0.81 ± 0.01 \\
Immature vs mature lymphocyte & 65 vs. 26,242 & \textbf{0.98 ± 0.00} & 0.89 ± 0.02 \\
Faggot cell vs promyelocyte & 47 vs. 11,994 & 0.91 ± 0.03 & \textbf{0.92 ± 0.01} \\
\bottomrule
\end{tabular*}\fi
\end{table}

\subsection{Skin axes against dermatologists' annotations}\label{sec:S9.4}

\paragraph{Axis contrasts (main Sec. 5.3).} Five of the six skin axes follow the steps of dermoscopic diagnosis (Table \hyperref[tab:S4]{S4}). E1 makes the first step of the two-step algorithm \cite{Marghoob2010}, as it opposes the melanocytic melanoma and nevus to the keratinocytic benign and actinic keratoses. E2, E3, and E5 separate diagnoses by the criteria of the second step. E2 opposes nevus and melanoma to vascular lesions and dermatofibroma through pigment network, E3 opposes basal-cell carcinoma and actinic keratosis to benign keratosis and melanoma through ulceration and blue-gray ovoid nests, and E5 opposes basal-cell carcinoma and vascular lesions to actinic keratosis and dermatofibroma through red-blue lacunae. E4 opposes dermatofibroma to actinic keratosis by lesion tone. The sixth axis, E0, separates excised lesions from follow-up lesions. Within each class, the share of histopathology-confirmed lesions drops by 1.58 s.d. toward its + pole, so E0 largely tracks how a diagnosis was obtained.

\paragraph{Name check.} MONET names each pole of the skin axes, and the Task 2 masks, which mark where dermatologists saw five structures on 2,590 lesions, show whether a name is right. For each axis, we take the 20\% of nevi, melanomas, and benign keratoses furthest toward each pole and ask which end holds more lesions with a marked structure. A permutation test within each class, with Benjamini–Hochberg correction \cite{Benjamini1995} over the 30 pairs of axis and structure, gives a clear answer for 17 pairs (FDR \textless{} 0.05). Each of the five structures is among them at least twice, so the masks place every structure on the axes. A name is checkable when MONET names a pole with one of the five structures on an axis where the masks give an answer. Four of the five checkable names match the marked structures (Table \hyperref[tab:S3]{S3}). The names come from the classification images alone, and this name check is out of sample. The matches include the two clearest cases, where the share of lesions with a marked pigment network is 0.37 higher toward the − pole of E1 and 0.40 higher toward the + pole of E5, while the one mismatch is on the acquisition axis E0. All five PBC names checked against maturation stages agree.

\paragraph{Structure scores.} Within each class, a lower E1 position marks lesions with pigment network, globules, and streaks (within-class AUROC 0.68 ± 0.00, 0.63 ± 0.01, and 0.69 ± 0.02 over seeds), and of the six axes E1 separates each of these three structures best. Milia-like cysts lie toward the other end. Within melanoma and nevus, E1 separates lesions with one of the three structures and no milia-like cyst from lesions with milia-like cysts and none of the three (AUROC 0.75 ± 0.02, and 0.82 ± 0.01 over all diagnoses). From the six axis positions alone, the five structures reach a mean held-out AUROC of 0.69 ± 0.01, against 0.58 from the diagnosis alone, and each is predicted better (Table \hyperref[tab:S3a]{S3a}).

\begin{table}[tbp]
\centering
\phantomsection\label{tab:S3a}\caption*{\textbf{Table S3a. The six skin axis positions predict the dermatologists' structures better than the diagnosis.} Held-out AUROC on the Task 2 test lesions, mean ± s.d. over three seeds.}
\footnotesize
\setlength{\tabcolsep}{4pt}
\sbox\fittab{\begin{tabular}{@{}lccc@{}}
\toprule
Structure & Diagnosis alone & Six axis positions alone & Gain \\
\midrule
Pigment network & 0.61 & 0.78 ± 0.01 & +0.17 \\
Globules & 0.51 & 0.64 ± 0.01 & +0.13 \\
Streaks & 0.55 & 0.68 ± 0.05 & +0.13 \\
Negative network & 0.69 & 0.78 ± 0.01 & +0.09 \\
Milia-like cysts & 0.54 & 0.55 ± 0.00 & +0.01 \\
Mean & 0.58 & 0.69 ± 0.01 & +0.11 \\
\bottomrule
\end{tabular}}%
\ifdim\wd\fittab>\columnwidth\resizebox{\columnwidth}{!}{\usebox\fittab}\else
\begin{tabular*}{\columnwidth}{@{\extracolsep{\fill}}lccc}
\toprule
Structure & Diagnosis alone & Six axis positions alone & Gain \\
\midrule
Pigment network & 0.61 & 0.78 ± 0.01 & +0.17 \\
Globules & 0.51 & 0.64 ± 0.01 & +0.13 \\
Streaks & 0.55 & 0.68 ± 0.05 & +0.13 \\
Negative network & 0.69 & 0.78 ± 0.01 & +0.09 \\
Milia-like cysts & 0.54 & 0.55 ± 0.00 & +0.01 \\
Mean & 0.58 & 0.69 ± 0.01 & +0.11 \\
\bottomrule
\end{tabular*}\fi
\end{table}

\subsection{Errors and oracle interventions}\label{sec:S9.5}

\paragraph{Error margin (main Sec. 5.4).} PBC errors are pooled over the validation and test images of the three seeds. In all 96 errors, E0 carries most of the error margin, 58\% on average (57\% for neutrophils called immature granulocytes and 62\% for the reverse), against 30\% for the next axis, E1.

\paragraph{Oracle correction.} Following the test-time interventions of concept bottlenecks \cite{CBM}, we take, for each image, the axis that argues most against the true class, replace its evidence by the mean training evidence of the true class, and recompute the logits. Concepts are corrected the same way, worst first. One corrected axis does the work of 12 corrected concepts on skin and BMC, and 24 on PBC and OBM (Table \hyperref[tab:S13]{S13}).

\begin{table}[tbp]
\centering
\phantomsection\label{tab:S13}\caption*{\textbf{Table S13. One corrected axis does the work of 12 to 24 corrected concepts.} Validation balanced accuracy, mean of 3 seeds, worst parts corrected first. The last column is the fewest corrected concepts that reach the accuracy of one corrected axis.}
\footnotesize
\setlength{\tabcolsep}{4pt}
\sbox\fittab{\begin{tabular}{@{}lccccc@{}}
\toprule
Dataset & \makecell{None} & \makecell{1 axis} & \makecell{1 concept} & \makecell{24 concepts} & \makecell{Concepts\\matching 1\\axis} \\
\midrule
Skin & 82.3 & 94.8 & 85.1 & 96.9 & 12 \\
PBC & 99.2 & 99.7 & 99.3 & 99.8 & 24 \\
OBM & 75.3 & 83.2 & 78.0 & 83.8 & 24 \\
BMC & 83.9 & 93.2 & 85.6 & 96.1 & 12 \\
\bottomrule
\end{tabular}}%
\ifdim\wd\fittab>\columnwidth\resizebox{\columnwidth}{!}{\usebox\fittab}\else
\begin{tabular*}{\columnwidth}{@{\extracolsep{\fill}}lccccc}
\toprule
Dataset & \makecell{None} & \makecell{1 axis} & \makecell{1 concept} & \makecell{24 concepts} & \makecell{Concepts\\matching 1\\axis} \\
\midrule
Skin & 82.3 & 94.8 & 85.1 & 96.9 & 12 \\
PBC & 99.2 & 99.7 & 99.3 & 99.8 & 24 \\
OBM & 75.3 & 83.2 & 78.0 & 83.8 & 24 \\
BMC & 83.9 & 93.2 & 85.6 & 96.1 & 12 \\
\bottomrule
\end{tabular*}\fi
\end{table}

\subsection{Planted shortcut}\label{sec:S9.6}

\paragraph{Protocol (main Sec. 5.4).} The dot is drawn at a random position, with radius 10 px at 224 px for the strong setting and 6 px, under half a patch, for the weak one. To locate the dot in the basis, we show each image with and without the dot and record how much the dot raises the melanoma evidence, the melanoma logit over the mean of the other classes. Because the logits decompose exactly over the axes, this rise splits into one part per axis, and an axis's share of the dot is its part of the rise. On the marker axis nearly all of this part lies on one pole, as the other pole holds under 4\% of the rise. Validation images choose the marker axis and pole, and test images give the shares of Table \hyperref[tab:S14]{S14}. Removing an axis sets its evidence to its training mean, with class offsets recalibrated on validation. Unlike the pole names of main Sec. 3.4, this audit names only the marker pole, by contrasting the test images furthest toward it with the ones furthest toward the opposite pole. As a dot is not a dermoscopic structure, the audit uses MedSigLIP, not MONET, with 20 clinical terms (Table \hyperref[tab:S1]{S1} and dotted vessels), 8 artifact terms, and dots of six colors, five of them absent from the images. MedSigLIP ranks ``a green dot'' first in all three seeds, ahead of every clinical term and every other color.

\begin{table}[tbp]
\centering
\phantomsection\label{tab:S14}\caption*{\textbf{Table S14. A planted dot lands on one axis at both strengths and is removed with it.} Skin, test, mean over 3 seeds. Rows 2–5 are changes in balanced accuracy (points). One model per dot, the strong dot on 90\% of melanomas and 5\% of other lesions, the weak dot on 80\% and 10\%. Reversed moves the dot onto the other classes. A dot-free model scores 84.6 aligned and 84.9 reversed.}
\footnotesize
\setlength{\tabcolsep}{4pt}
\sbox\fittab{\begin{tabular}{@{}lcc@{}}
\toprule
Training marker & \makecell{Strong dot} & \makecell{Weak dot} \\
\midrule
Dot share on one pole & 0.83 & 0.74 \\
Aligned → reversed & −25.1 & −11.5 \\
Axis removal, reversed & +19.9 & +5.2 \\
Axis removal, unmarked & +1.6 & +1.2 \\
Best other-axis removal & −1.2 & −0.6 \\
\bottomrule
\end{tabular}}%
\ifdim\wd\fittab>\columnwidth\resizebox{\columnwidth}{!}{\usebox\fittab}\else
\begin{tabular*}{\columnwidth}{@{\extracolsep{\fill}}lcc}
\toprule
Training marker & \makecell{Strong dot} & \makecell{Weak dot} \\
\midrule
Dot share on one pole & 0.83 & 0.74 \\
Aligned → reversed & −25.1 & −11.5 \\
Axis removal, reversed & +19.9 & +5.2 \\
Axis removal, unmarked & +1.6 & +1.2 \\
Best other-axis removal & −1.2 & −0.6 \\
\bottomrule
\end{tabular*}\fi
\end{table}

\paragraph{The dot across concepts.} Over the 144 concepts, 80\% of the dot's evidence needs 12 concepts from all twelve layers (the strongest holds 8.5 to 9.7\%), and two seeds share few of them (Jaccard 0.21 to 0.33).

\subsection{Acquisition shortcuts}\label{sec:S9.7}

\paragraph{Flagged axes (main Sec. 5.4).} The five flagged axes are skin E0 (excised vs follow-up), BMC E2 (crowding), and chest E0, E2, and E4 (view position and film rotation, Table \hyperref[tab:S4]{S4}).

\begin{table}[tbp]
\centering
\phantomsection\label{tab:S15}\caption*{\textbf{Table S15. Acquisition audit.} Test, mean over 3 seeds, gap CI from the bootstrap of the text. Var. AUROC is how well the position separates the groups. * Not via the film corners.}
\footnotesize
\setlength{\tabcolsep}{4pt}
\sbox\fittab{\begin{tabular}{@{}lccll@{}}
\toprule
Axis (variable) & \makecell{Var. AUROC} & \makecell{AUROC with\\/ against} & Gap (95\% CI) & Verdict \\
\midrule
Chest E0, E2 (view) & 0.85 / 0.86 & 0.881 / 0.656 & 0.225 (0.210–0.242) & relied on \\
Chest E4 (rotation) & 0.79 & 0.821 / 0.775 & 0.046 (0.030–0.062) & small * \\
BMC E2 (crowding) & 0.74 & 0.984 / 0.977 & 0.007 (0.004–0.010) & negligible \\
Skin E0 (excision) & 0.86 & 0.983 / 0.963 & 0.020 (0.003–0.037) & small \\
\bottomrule
\end{tabular}}%
\ifdim\wd\fittab>\columnwidth\resizebox{\columnwidth}{!}{\usebox\fittab}\else
\begin{tabular*}{\columnwidth}{@{\extracolsep{\fill}}lccll}
\toprule
Axis (variable) & \makecell{Var. AUROC} & \makecell{AUROC with\\/ against} & Gap (95\% CI) & Verdict \\
\midrule
Chest E0, E2 (view) & 0.85 / 0.86 & 0.881 / 0.656 & 0.225 (0.210–0.242) & relied on \\
Chest E4 (rotation) & 0.79 & 0.821 / 0.775 & 0.046 (0.030–0.062) & small * \\
BMC E2 (crowding) & 0.74 & 0.984 / 0.977 & 0.007 (0.004–0.010) & negligible \\
Skin E0 (excision) & 0.86 & 0.983 / 0.963 & 0.020 (0.003–0.037) & small \\
\bottomrule
\end{tabular*}\fi
\end{table}

\paragraph{Reliance test.} On test, we pair each class's positives from the group where the variable makes the class rare with negatives from the group where it is common (against the prior), and the reverse (with the prior). Groups are the two view positions, or the outer quartiles of a continuous variable, with class priors taken from validation. The with-minus-against AUROC gap measures how much the ranking depends on the variable, with 95\% CIs from a bootstrap over patients (chest), lesions (skin), or images (BMC) (Table \hyperref[tab:S15]{S15}). View position is relied on (gap 0.225, CI 0.210 to 0.242). Every other gap stays below the reliance bar of 0.05, fixed before any result was seen, in all three seeds (0.046, 0.020, and 0.007), so by this test the classifier does not rely on them. As they are not exactly zero, we also test film rotation causally, the largest gap and the one cue we can change while holding the anatomy. Filling the black film corners with the film's border intensity removes them (4.6\% to 0.2\% of the film) but leaves the gap in place (0.046 to 0.051). Like the planted dot (Sec. \hyperref[sec:S9.6]{S9.6}), this causal test shows the classifier does not read the corners, and the remaining gap comes from other content of these films.